\documentclass[11pt]{article}

\usepackage[final]{acl}
\usepackage{times}
\usepackage{bbding}
\usepackage{latexsym}
\usepackage[T1]{fontenc}
\usepackage[utf8]{inputenc}
\usepackage{microtype}
\usepackage{graphicx}
\usepackage{amsmath,amssymb}
\usepackage{enumitem}
\usepackage{listings}
\usepackage{microtype}
\usepackage{inconsolata}
\usepackage{graphicx}
\usepackage{xspace}
\usepackage[table]{xcolor}
\usepackage{todonotes}
\usepackage{booktabs}
\usepackage{multirow}
\usepackage{threeparttable}
\usepackage{adjustbox}
\usepackage{booktabs}
\usepackage[most]{tcolorbox}
\usepackage{arydshln}
\usepackage{float}
\usepackage[table,dvipsnames]{xcolor}
\usepackage{pgfplots}
\pgfplotsset{compat=1.18}
\usepgfplotslibrary{groupplots}

\newcommand{\eg}{\emph{e.g.,}\xspace}

\newcommand{\etal}{\emph{et~al.}\xspace}
\newcommand{\secref}[1]{Section~\ref{#1}\xspace}

\newcommand{\figref}[1]{Fig.~\ref{#1}\xspace}

\newcommand{\tabref}[1]{Table~\ref{#1}\xspace}
\newcommand{\qone}{\textbf{Q1}\xspace}
\newcommand{\qtwo}{\textbf{Q2}\xspace}
\newcommand{\qthree}{\textbf{Q3}\xspace}

\definecolor{biashlcolor}{HTML}{8B0000}    
\definecolor{ceilhlcolor}{HTML}{006400}    
\definecolor{relaccRed}{HTML}{B00020}
\newcommand{\biashl}[1]{\textcolor{biashlcolor}{\strut\textbf{#1}}}
\newcommand{\ceilhl}[1]{\textcolor{ceilhlcolor}{\strut\textbf{#1}}}
\newcommand{\relaccval}[1]{\textcolor{relaccRed}{#1}}

\newcommand{\qwen}[1]{\text{Qwen3-#1}}
\newcommand{\intern}[1]{\text{InternVL3.5-#1}}
\newcommand{\qwenfam}{\text{Qwen3}\xspace}
\newcommand{\internfam}{\text{InternVL3.5}\xspace}
\newcommand{\gptlarge}{\text{GPT-5.4}\xspace}
\newcommand{\gptmini}{\text{GPT-5.4 Mini}\xspace}
\newcommand{\gptfam}{\text{GPT-5.4}\xspace}\makeatletter
\def\adl@drawiv#1#2#3{%
        \hskip.5\tabcolsep
        \xleaders#3{#2.5\@tempdimb #1{1}#2.5\@tempdimb}%
                #2\z@ plus1fil minus1fil\relax
        \hskip.5\tabcolsep}
\newcommand{\cdashlinelr}[1]{%
  \noalign{\vskip\aboverulesep
           \global\let\@dashdrawstore\adl@draw
           \global\let\adl@draw\adl@drawiv}
  \cdashline{#1}
  \noalign{\global\let\adl@draw\@dashdrawstore
           \vskip\belowrulesep}}
\makeatother
\definecolor{color1}{RGB}{210, 235, 180}
\definecolor{color2}{RGB}{240, 190, 140}
\definecolor{color3}{RGB}{220, 200, 230}
\definecolor{color4}{RGB}{150, 170, 210}
\definecolor{brand}{HTML}{1A9D8F}      
\definecolor{brandD}{HTML}{2A7F6F}     
\definecolor{ink}{HTML}{1A1A1A}
\definecolor{keybg}{HTML}{E8F4F1}      
\definecolor{keydark}{HTML}{2F5E58}    
\definecolor{accent}{HTML}{FF6B6B}
\definecolor{accentD}{HTML}{D94F4F}    
\definecolor{soft}{HTML}{F3F8F7}
\definecolor{xhsRed}{HTML}{FF4D4F}
\definecolor{xhsDark}{HTML}{1F1F1F}
\definecolor{xhsCream}{HTML}{FFF8F0}
\definecolor{xhsBlue}{HTML}{2F54EB}
\definecolor{xhsYellow}{HTML}{FFB800}

\title{Prior Bias in Vision Language Models on UML Diagram Interpretation}

\author{
Zaiyu Cheng\thanks{Equal contribution.} \\
  Dept. of Computer Science\\
  William \& Mary, USA \\
  \texttt{zcheng06@wm.edu} \\\And
Khai-Nguyen Nguyen\footnotemark[1] \\
  Dept. of Computer Science\\
  William \& Mary, USA \\
  \texttt{nkn002@bucknell.edu} \\\And
Antonio Mastropaolo \\
  Dept. of Computer Science\\
  William \& Mary, USA \\
  \texttt{amastropaolo@wm.edu}
}

\begin{document}
\maketitle
\begin{abstract}
Vision Language Models (VLMs) are increasingly applied to software engineering artifacts, especially UML class diagrams whose meaning depends on visual notation. Yet, it is unclear whether VLMs actually read such diagrams or instead answer from pretrained priors about how classes typically relate. We introduce a controlled UML benchmark in which each prior-conforming diagram is paired with its prior-conflicting counterpart that (1) preserves the same class names and layout while (2) reverses only the relation arrow. We evaluate eight open-source VLMs from two model families, \internfam and \qwenfam, alongside two closed-source frontier models \gptlarge and \gptmini. Across the eight open-source models, reversing the arrow reduces relation-direction accuracy by \textbf{33.48\%} on average, while \gptmini retains a \textbf{10\%} gap. In the harder three-class condition, accuracy drops sharply by \textbf{45.28\%} for open-source models, and even \textbf{18.62\%} for the \gptlarge family on average. Scaling provides only limited improvements and is family-dependent. Our benchmark presents a diagnostic prior-driven failure in diagram-grounded software understanding. Our artifact is available at {\url{https://anonymous.4open.science/r/UMLKnowledgeConflict-8461}}.
\end{abstract}

\section{Introduction}
Vision Language Models (VLMs) inherit a vast amount of prior knowledge about how objects are related from their LLM backbones. While this prior knowledge can be useful, it is also a known source of failure: when visual evidence contradicts what the model already ``knows'', generation often follows the prior rather than the image~\cite{lin2023revisiting,li2023evaluating,guan2024hallusionbench,golovanevsky2025pixels,vo2025vision, conrardy2024image,antal2024toward}.

\begin{figure}[!ht]
    \centering
    \includegraphics[width=0.95\linewidth]{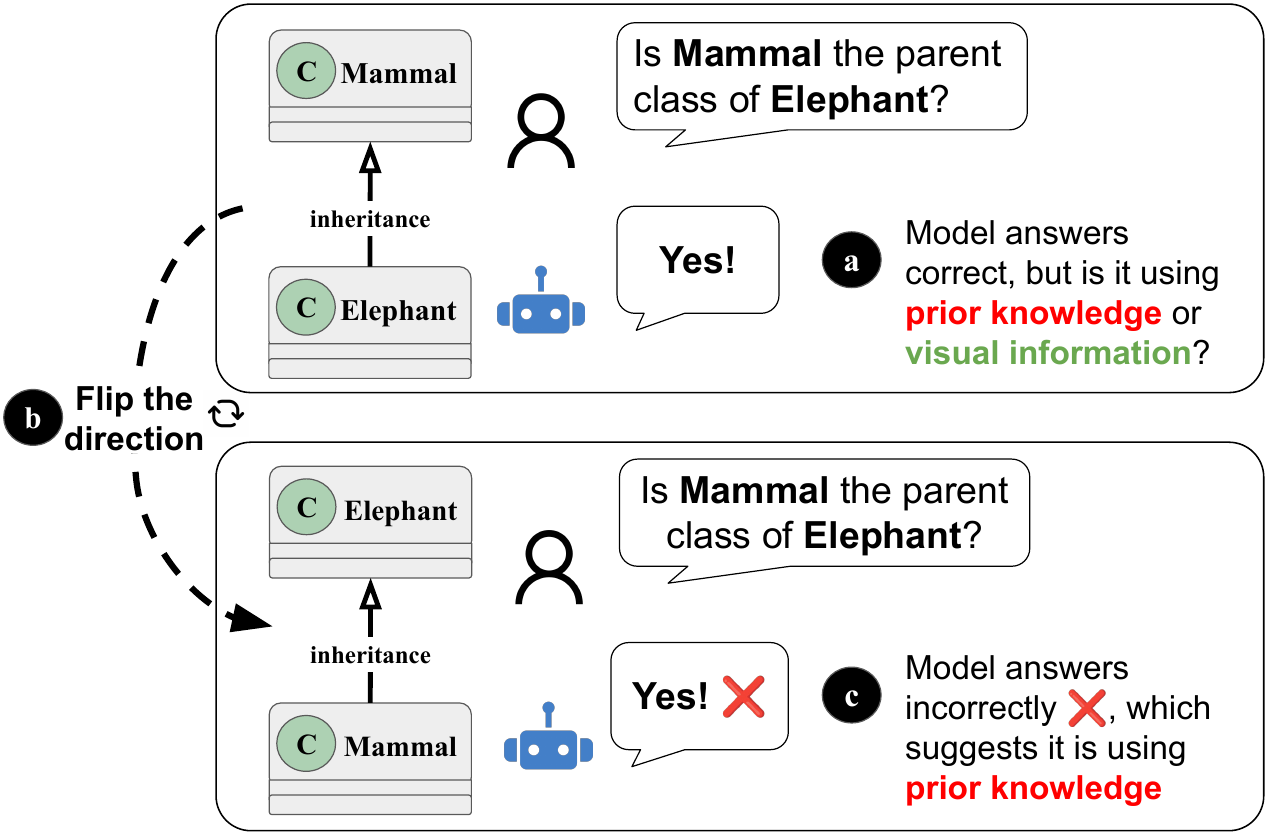}
    \caption{\textbf{VLMs follow the class names, not the arrow.} (a) On a prior-conforming diagram, VLMs correctly answer \emph{``Is Mammal the parent class of Elephant?''} with ``Yes.'' (b) Flipping the arrow makes ``No'' the correct answer. (c) VLMs still answer ``Yes'', following the prior knowledge rather than the depicted arrow.}
    \label{fig:example}
\end{figure}

Prior evidence~\cite{guan2024hallusionbench, liu2025phd, lee2025vlind, vo2025vision} that VLMs follow learned priors over visual evidence comes almost entirely from natural-image counterfactuals. Yet, VLMs are increasingly deployed in software-engineering pipelines that require reading visual artifacts whose meaning is determined solely by visual notation~\cite{zou2024vgbench, antal2024toward, hou2024vision}. It remains unclear how VLMs comprehend such diagrams, specifically how strongly their knowledge priors interfere when the diagram encodes a relation contradicting what the named entities encode in training. We aim to assess \textbf{whether VLMs interpret formal software diagrams from the visual notation or from the knowledge prior of the named entities}. For example, given a UML class diagram with an inheritance arrow from \texttt{Mammal} to \texttt{Elephant} (\figref{fig:example}), we ask ``Does \texttt{Elephant} inherit from \texttt{Mammal}?'', the correct answer is ``No'' because the arrow points the other way, but VLMs systematically answer ``Yes'' by following the knowledge prior.

To study this question, we narrow our focus specifically to UML diagrams, which expose this fragility cleanly. A UML class diagram is a \emph{formal specification} whose meaning is fixed by \textbf{visual notation}, not by what the classes are called~\cite{rumpe2016modeling}. We focus on four class relations represented by arrows with a direction that flips the relation if reversed: \textit{inheritance}, \textit{composition}, \textit{dependency}, and \textit{aggregation}. The same two class names can therefore encode either the prior-conforming or the prior-conflicting relation, depending only on which way the arrow points~\cite{roberts2024image2struct}.

To empirically study this failure mode, we introduce a controlled UML benchmark. Every prior-conflicting diagram is paired with a prior-conforming counterpart built from the same class names, the same notation, and only a flipped direction (\figref{fig:example})~\cite{wu2025lanp}. A further \emph{prior-free} setting replaces meaningful class names with opaque labels, separating knowledge priors from visual grounding~\cite{li2024naturalbench}. The benchmark covers the four direction-sensitive UML relations (inheritance, aggregation, composition, and dependency) and evaluates 8 open-source model sizes across two VLM families, InternVL3.5 \cite{wang2025internvl3} and Qwen3 \cite{bai2025qwen3}, plus two closed-source frontier models, \gptlarge and \gptmini. Our key findings are:

\begin{enumerate}
[leftmargin=*,itemsep=2pt,topsep=2pt]
\item \textbf{Knowledge prior dominates visual evidence in open-source VLMs, while closed-source models reduce but do not eliminate errors on harder settings.} Reversing the arrow collapses relation accuracy by \textbf{33.48\% on average} across the two open-source model families, with \internfam dropping more accuracy (-40.52 \%) than \qwenfam (-26.44 \%); \gptmini reduces the  two-class drop to $-10.00$\%, and \gptlarge nearly closes it ($\Delta \approx 0$). On harder three-class variants, accuracy even drops by \textbf{45.28\%} for open-source models, and \textbf{18.62\%} for the \gptlarge family (\secref{sec:finding1}).

\item \textbf{Adding more visual context exacerbates the prior-over-vision failure mode.} Specifically, adding one auxiliary edge to the prior-conflicting UML diagrams further reduces accuracy by $\sim$12\% on average. Adding more details increases visual complexity, leading to worse model performance (\secref{sec:finding1}).
\item \textbf{The prior-over-vision effect is robust across image scales, and is focused on Aggregation and Inheritance.} On three image scales of $1\times$, $1.5\times$, $2\times$, the collapse is consistent, confirming it is not an artifact of a perceptual bottleneck. On four tested relations, models exhibit the prior-over-vision effect clearly on Aggregation and Inheritance (\eg $\sim50\%$ accuracy loss when flipping the arrow) (\secref{sec:finding2}).
\item \textbf{The effect of model scaling is governed by model visual capabilities.} In Qwen3, increasing the model size helps on two-class prior-conflicting UML diagrams (+58.55\% from 2B to 32B). On the same subset, InternVL3.5 with 2B parameters uniquely outperforms all larger variants, but this is due to the weak visual processing capabilities of InternVL3.5 (\secref{sec:finding3}).
\end{enumerate}


\section{Related Work}
\label{sec:related_work}

\subsection{UML Interpretation by Models}
\label{sub:uml_understanding}
Earlier work on UML class-diagram understanding built dedicated extraction pipelines that recover structured semantics directly from images. Hammond and Davis~\cite{hammond2006tahuti} introduced \textit{Tahuti}, a sketch-based recognizer combining stroke-level parsing with interactive correction; Lank \etal~\cite{lank2000interactive} proposed pen-based UML interpretation with user-in-the-loop disambiguation; Karasneh and Chaudron~\cite{karasneh2013img2uml,karasneh2013online} presented \textit{Img2UML} and released a large web-mined UML repository; and Chen \etal~\cite{chen2022automatically} learned semantic element recognition with \textit{ReSECDI}. More recent work tests whether end-to-end multimodal models can read UML directly without such pipelines~\cite{de2024evaluating}: Antal \etal~\cite{antal2024toward} report that GPT-4V handles simple class diagrams but degrades on complex relation structures; Naboichenko and Peinl~\cite{naboichenko2026unlocking} find that VLMs still struggle with relation-centric questions over UML class diagrams; Shehzadi \etal~\cite{shehzadi2025automatic} release a UML-focused VQA dataset; Wang \etal~\cite{wang2025assessing} investigate GPT-based grading of student UML submissions; and Campanello \etal~\cite{campanello2025use} analyze LLM-supported UML reverse-engineering. These studies establish that VLMs can read UML labels but report uneven relation accuracy. They do not, however, isolate whether the failure is perceptual or driven by pretrained priors about how the named classes typically relate.

\begin{figure*}[!ht]
    \centering
    \includegraphics[width=0.92\linewidth]{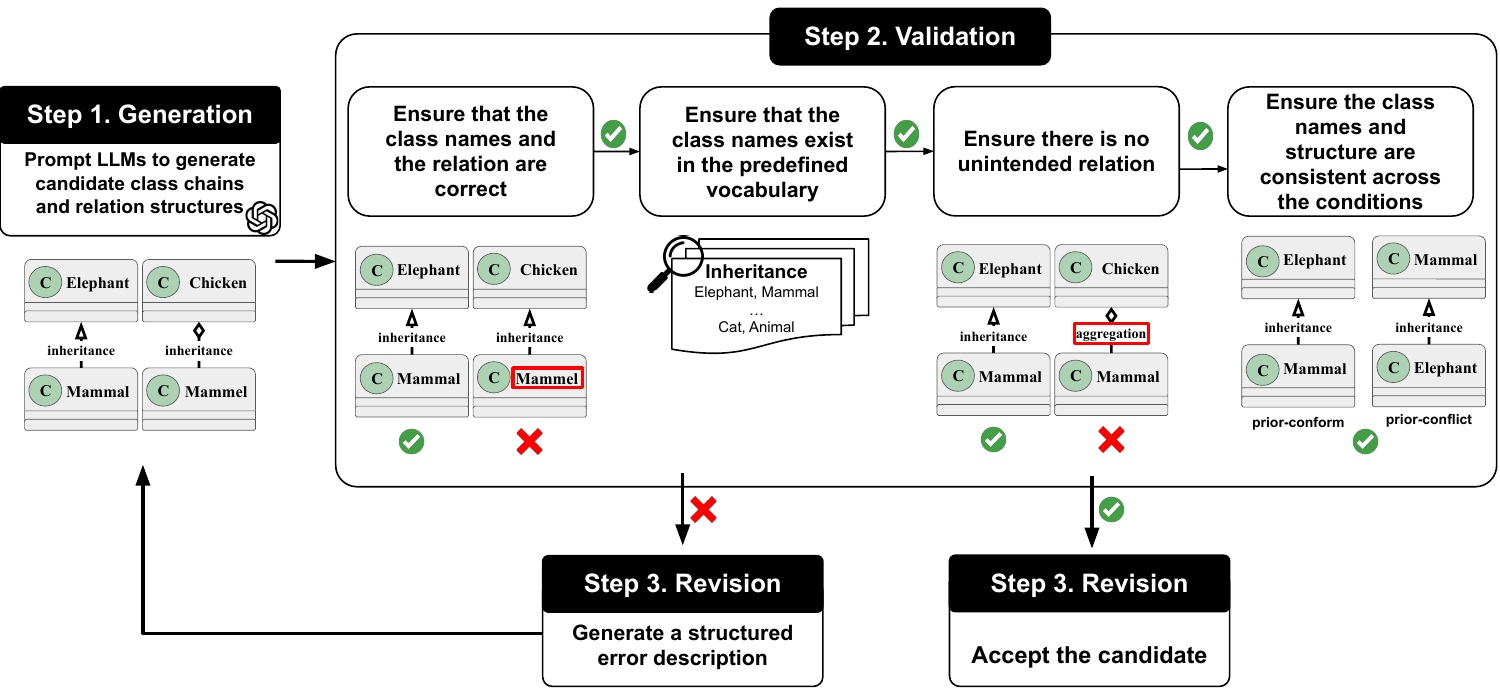}
    \caption{\textbf{Data generation pipeline.} (1) We prompt SOTA LLMs to generate candidate UML diagrams across the three conditions; (2) We then validate the diagrams against four constraints, ensuring consistency and correctness of these images; (3) If there is any error (\textcolor{red}{\XSolidBold}), we generate a structured error description and redo step (1), else (\textcolor{OliveGreen}{\CheckmarkBold}) we accept the candidate.}
    \label{fig:generation_pipeline}
\end{figure*}

\subsection{Hallucination and Prior Bias in VLMs}
\label{sub:vlm_priors}
A growing line of work documents that VLMs produce fluent-but-unfaithful answers, especially on images that violate common-sense expectations~\cite{liu2024survey}. Early VQA studies showed that strong language priors can dominate image evidence~\cite{goyal2017making,ramakrishnan2018overcoming,wu2022overcoming}; subsequent benchmarks measure this directly through object hallucination~\cite{li2023evaluating,huang2024visual}, entangled language and visual illusion~\cite{guan2024hallusionbench}, and counter-commonsense images~\cite{liu2025phd,bitton2023breaking}. These works share a methodological lesson: aggregate accuracy obscures the locus of failure, and disaggregating perception from inference is essential~\cite{zhu2025benchmarking}. A more recent thread isolates the vision--knowledge conflict directly~\cite{li2024visiongraph}. \textit{VLind-Bench}~\cite{lee2025vlind} measures how strongly language priors override visual input; \textit{ViLP}~\cite{luo2024probing} probes whether VLMs trust pixels over learned facts; and Liu \etal~\cite{liu2024insight} report that ``insight over sight'' bias persists across modern multimodal LLMs. Most recently, Vo \etal~\cite{vo2025vision} demonstrate that current VLMs systematically follow common-sense priors over visual evidence on counterfactual images of natural objects. Our work extends this line to formal software diagrams, where the source of truth is notation rather than pixel-level photorealism, and where the prior conflict is constructible by reversing a single arrow rather than relying on naturally-occurring counterfactuals.
\section{Benchmark and Protocol}

\subsection{Benchmark Construction}
\label{sec:construction}
Our benchmark is designed to isolate prior-driven failure from generic task difficulty. For example, let ``$\rightarrow$'' denote the ``inherits from'' relation, every prior-conforming UML diagram (\eg \texttt{Dog} $\rightarrow$ \texttt{Animal}) is paired with a prior-conflicting counterpart (\eg \texttt{Animal} $\rightarrow$ \texttt{Dog}). The two diagrams in a pair share class names, layout, and queries; they differ only in the direction of the relationship arrow.

\paragraph{Instances.}
Each benchmark instance is a tuple $x = (I, r, (c_i, c_j), y)$, where $I$ is the rendered UML class-diagram image, $r \in \{$\textit{inheritance, aggregation, composition, dependency}$\}$ is the queried relation type, $(c_i, c_j)$ is an ordered pair of class names appearing in $I$, and $y \in \{$True, False$\}$ is the gold label. The query has the form ``In this diagram, does the relation $r$ hold from $c_i$ to $c_j$?'', and the model must respond with one label from $\{$True, False, Unknown$\}$; the Unknown option lets the model abstain on genuinely unreadable diagrams rather than be forced to guess, but abstentions count as incorrect during scoring. The four relations are rendered in standard UML notation: \textit{inheritance} as a hollow triangle pointing toward the parent class, \textit{aggregation} as a hollow diamond, \textit{composition} as a filled diamond, and \textit{dependency} as a dashed arrow pointing toward the depended-on class.

\begin{figure}[!ht]
    \centering
    \includegraphics[width=0.95\linewidth]{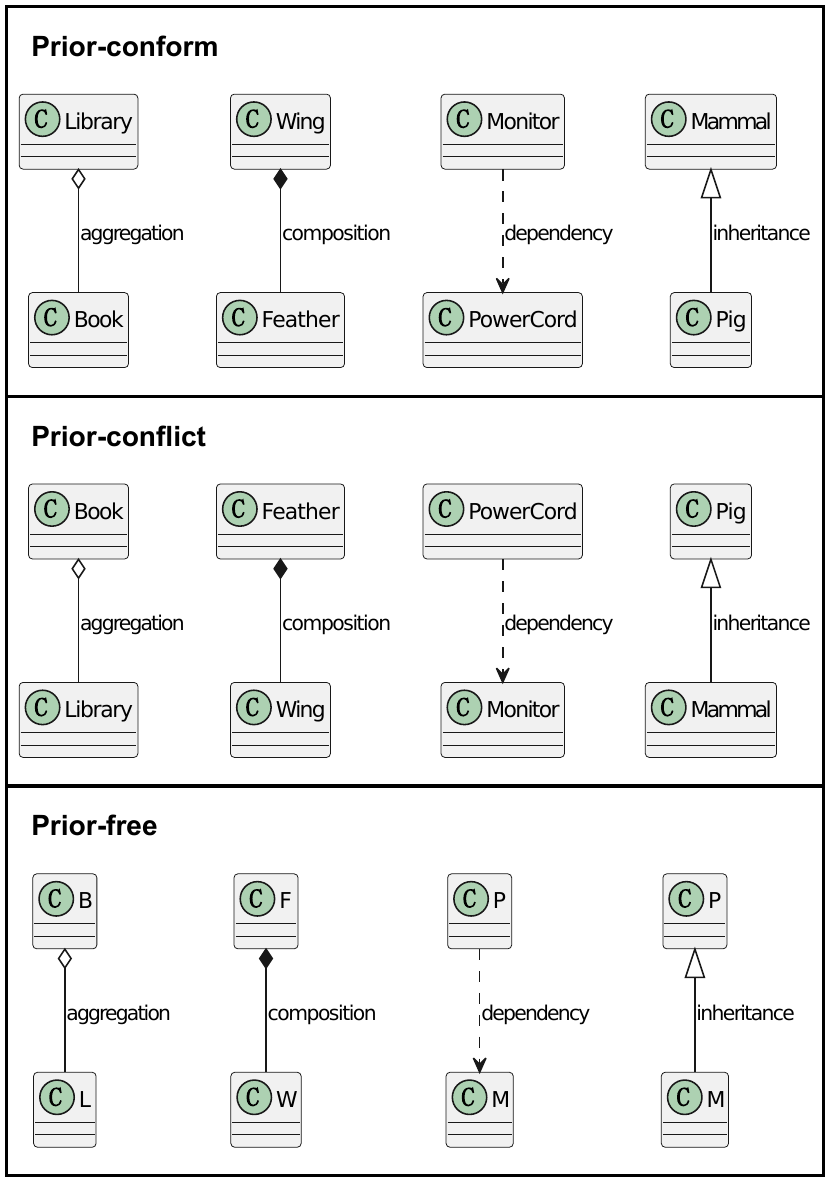}
    \caption{\textbf{Examples for each condition type.} We report only the \textit{2-reverse} subset for the prior-conflict condition. Additional illustrations, including other prior-conflict subsets, are shown in \figref{fig:full_examples}.}
    \label{fig:conditions}
\end{figure}

\paragraph{Conditions.}
Our hypothesis is that VLMs follow the knowledge prior implied by the class names rather than the depicted relation arrow. Following this hypothesis, we design three conditions that contrast prior alignment against visual evidence while holding diagram structure (number of classes, edge type, layout) fixed:
\begin{enumerate}[leftmargin=*,itemsep=2pt,topsep=2pt]
  \item \textbf{Prior-conforming.} The visual evidence and knowledge prior point to the same answer. We design the relation to follow canonical facts induced by the class names (\eg \texttt{Dog} $\rightarrow$ \texttt{Animal}).
  \item \textbf{Prior-conflicting.} The visual evidence and the prior  point to opposite answers. We use the same class names and relation type as the prior-conforming, but reverse the depicted relation so the image contradicts canonical expectations. To study knowledge conflict, we include (i) \textit{2-reverse}, the standard two-class condition (\eg \texttt{Animal} $\rightarrow$ \texttt{Dog}), (ii) \textit{3-reverse}, a three-class condition where all pairs are prior-conflicting (\eg \texttt{Animal} $\rightarrow$ \texttt{Mammal} $\rightarrow$ \texttt{Dog}), and (iii) \textit{3-mixed}, a three-class condition where one pair is prior-conforming and the other is prior-conflicting (\eg \texttt{Animal} $\leftarrow$ \texttt{Mammal} $\rightarrow$ \texttt{Dog}).
 
  \item \textbf{Prior-free.} No knowledge prior is activated, and the model must use purely the visual information. We use the initial letters from the prior-conforming labels to construct these samples (\eg \texttt{D} $\rightarrow$ \texttt{A} instead of \texttt{Dog} $\rightarrow$ \texttt{Animal}).
\end{enumerate}

The matched \textit{prior-conforming} vs.\ \textit{prior-conflicting} contrast measures \emph{how much} a model's answer changes when visual evidence contradicts the knowledge prior. But the model may fail for two reasons: (1) bias toward the prior, or (2) inability to read the diagram in the first place. The \textit{prior-free} condition disentangles these two failure modes: if the model fails on \textit{prior-free} (where no prior knowledge is available), the failure on \textit{prior-conflicting} is consistent with (2) and not bias~\cite{fu2024blink}.

\paragraph{Generation.}
We develop the benchmark with an LLM-assisted pipeline consisting of three steps illustrated in \figref{fig:generation_pipeline}. We detail the steps as follows:
\begin{enumerate}
    \item \textbf{Generate candidates.} High-capability LLMs are prompted to generate candidate class chains and relation structures consistent with the requested condition.
    \item \textbf{Validate candidates.} Each candidate is validated against four constraints: (i) the queried relation $r$ and ordered pair $(c_i, c_j)$ exist in the candidate as specified; (ii) class names are drawn from a controlled vocabulary chosen so the prior is unambiguous; (iii) no other unintended relation of type $r$ appears elsewhere in the diagram; and (iv) for matched variants, class names and edge structure agree with the matched counterpart.
    \item \textbf{Revise candidates.} Candidates that fail any check are returned to the LLM with a structured error description and revised. The repair loop continues until every accepted candidate passes all four checks.
\end{enumerate}

\subsection{Evaluation Protocol}
\label{sec:protocol}
A UML failure can be of two kinds: a model may fail to read the class names at all, or read them perfectly yet still answer the relation question incorrectly. We separate these failure modes with two metrics, one for perception and one for inference.

Let $\mathcal{D}$ be the evaluation set, $C(x)$ the gold class-name set for instance $x$, $\hat{C}(x)$ the model's predicted class-name set, $\mathcal{N}$ a deterministic name-normalization function (case folding, punctuation cleanup, alias normalization), $y(x)$ the gold relation label, and $\hat{y}(x)$ the predicted relation label. We define the two metrics as follows:

\paragraph{Class-name recovery.} Given only the diagram image, the model must output the complete set of class names appearing in the diagram (\eg \{\texttt{Dog}, \texttt{Animal}\}). This metric measures perception: whether the model can read the text labels in the diagram and outputs the exact match (we abbreviate this as EM).
$$\mathrm{EM} = \tfrac{1}{|\mathcal{D}|}\sum_{x \in \mathcal{D}} \mathbf{1}\big[\mathcal{N}(\hat{C}(x)) = \mathcal{N}(C(x))\big]$$

\paragraph{Relation-direction accuracy.} The model receives the same diagram image and a query about a specific ordered class pair (\eg ``In this diagram, does \texttt{Dog} inherit from \texttt{Animal}?'') and must respond with one label from $\{$True, False, Unknown$\}$. This metric measures whether the model can correctly ground the queried relation in the diagram's structure. We abbreviate this as \textit{RelAcc}.

$$\mathrm{RelAcc} = \tfrac{1}{|\mathcal{D}|}\sum_{x \in \mathcal{D}} \mathbf{1}\big[\hat{y}(x) = y(x)\big]$$

Our key effect-size metric is the \textbf{conflict gap}, denoted $\Delta$: the drop in accuracy when the depicted relation contradicts the knowledge prior:
\[
\Delta = \mathrm{RelAcc}_\textit{prior-conform} - \mathrm{RelAcc}_\textit{prior-conflict}
\]
A model unaffected by the conflict has $\Delta \approx 0$; a model fully dependent on the knowledge prior has $\Delta$ approaching its $\mathrm{RelAcc}_\textit{prior-conform}$. We report $\Delta$ both per-model and as a grand mean across models.

\section{Experimental Results}
\label{sec:results}

We organize the experiments around three research questions. \qone establishes the headline prior-override effect (\secref{sec:finding1}). \qtwo confirms that the effect is robust across the two axes built into the benchmark, relation type and image scale (\secref{sec:finding2}). \qthree studies whether scaling fixes the effect, and finds that results are dependent on visual capabilities and model family (\secref{sec:finding3}).

\subsection{\qone: Do VLMs follow class names rather than the arrow direction?}
\label{sec:finding1}
\subsubsection{Motivation}
\qone asks whether VLMs interpret a UML diagram from the depicted arrow or from the knowledge prior of the named entities. Our hypothesis is that priors dominate: when the arrow contradicts the prior (\eg ``Animal'' $\rightarrow$``Dog''), the model follows the prior rather than the arrow. We test this with the matched \textit{prior-conform} / \textit{prior-conflict} contrast (\figref{fig:example}), then extend to three-class conditions to ask whether adding more diagram context recovers visual grounding.

\subsubsection{Experiments}
We report $\mathrm{RelAcc}$ on \textit{prior-conform} and \textit{prior-conflict} images, averaged across four relation types and three image sizes. \figref{fig:finding1_conflict} reports the per-family breakdown with the conflict gap $\Delta$. We additionally compute the 10-model mean $\mathrm{RelAcc}$ on \textit{3-reverse} and \textit{3-mixed} (\figref{fig:finding1_three_class}) to measure how added structural context affects the accuracy.

\subsubsection{Results}
\paragraph{VLMs are mostly correct on prior-conforming diagrams but fail when the arrow is reversed.}
\figref{fig:finding1_conflict} shows a sharp drop from \textit{prior-conform} to \textit{prior-conflict}. The 10-model mean $\mathrm{RelAcc}$ falls from 74.14\% to 46.39\%, a knowledge conflict gap of \biashl{$\Delta = 27.75$ \%}; the drop is concentrated in the open-source families, whose 8-model mean shows a much larger gap of $\Delta = 33.48$\%. Both open-source families exhibit the same pattern: \qwenfam drops $79.04 \to 52.60$ ($\Delta = 26.44$); \internfam drops $59.15 \to 18.63$ ($\Delta = 40.52$). \internfam's gap is 14\% larger although the family starts lower on \textit{prior-conform} (a 59.15\% ceiling vs.\ Qwen's 79.04\%). This illustrates that strong prior knowledge (\eg high $\mathrm{RelAcc}_{\textit{conform}}$) does not always lead to a more pronounced prior-over-vision effect.

\begin{figure}[t]
\centering
\begin{tikzpicture}
\begin{axis}[
    width=\columnwidth,
    height=4.5cm,
    ybar,
    bar width=0.28cm,
    enlarge x limits=0.18,
    ymin=0, ymax=110,
    ytick={0,25,50,75,100},
    ylabel={RelAcc / $\Delta$ (\%)},
    symbolic x coords={Qwen3, InternVL3.5, GPT-5.4, Mean},
    xtick=data,
    label style={font=\footnotesize},
    tick label style={font=\footnotesize},
    grid=major,
    grid style={dashed, gray!55},
    legend style={font=\scriptsize, draw=gray!50, fill=white, at={(0.5,-0.22)}, anchor=north, legend columns=3, /tikz/every even column/.append style={column sep=0.4cm}},
    legend cell align=left,
    nodes near coords,
    nodes near coords style={font=\tiny, rotate=0, anchor=center, yshift=3pt, /pgf/number format/.cd, fixed, fixed zerofill, precision=0},
    tick align=inside
]
\addplot[fill=blue!45, draw=blue!80!black] coordinates {
  (Qwen3, 79.04) (InternVL3.5, 59.15) (GPT-5.4, 94.34) (Mean, 74.14)
};
\addlegendentry{\textit{prior-conform}}
\addplot[fill=orange!55, draw=orange!75!black] coordinates {
  (Qwen3, 52.60) (InternVL3.5, 18.63) (GPT-5.4, 89.52) (Mean, 46.39)
};
\addlegendentry{\textit{prior-conflict}}
\addplot[fill=biashlcolor!45, draw=biashlcolor!80!black] coordinates {
  (Qwen3, 26.44) (InternVL3.5, 40.52) (GPT-5.4, 4.82) (Mean, 27.75)
};
\addlegendentry{$\Delta$}
\end{axis}
\end{tikzpicture}
\caption{\textbf{Reversing the direction of a \textit{prior-conform} diagram lowers accuracy by 27.75\% on the all-model mean, but the drop is concentrated in the open-source families (\qwenfam $\Delta=26.44\%$, \internfam $\Delta=40.52\%$) while the \gptfam family makes the two-class gap smaller ($\Delta=4.82\%$).} Per-family $\mathrm{RelAcc}$ (\%) on \textit{prior-conform} (blue) and \textit{prior-conflict} (orange), with the conflict gap $\Delta$ (red); each family bar is the simple mean over its members (4 for \qwenfam and \internfam, 2 for \gptfam), and ``Mean'' is the simple mean over all 10 models.}
\label{fig:finding1_conflict}
\end{figure}
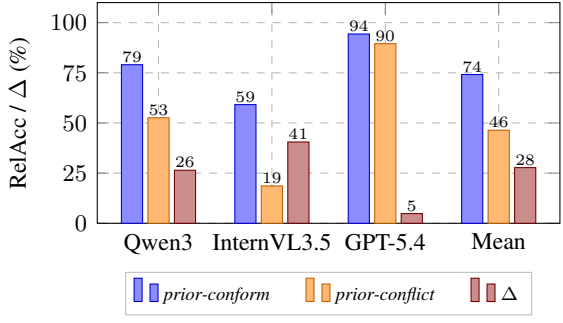
\begin{figure}[!ht]
\centering
\begin{tikzpicture}
\begin{axis}[
    width=\columnwidth,
    height=4.5cm,
    ybar,
    bar width=0.28cm,
    enlarge x limits=0.18,
    ymin=0, ymax=110,
    ytick={0,25,50,75,100},
    ylabel={$\mathrm{RelAcc}$ (\%)},
    symbolic x coords={Qwen3, InternVL3.5, GPT-5.4, Mean},
    xtick=data,
    label style={font=\footnotesize},
    tick label style={font=\footnotesize},
    grid=major,
    grid style={dashed, gray!55},
    legend style={font=\scriptsize, draw=gray!50, fill=white, at={(0.5,-0.22)}, anchor=north, legend columns=3, /tikz/every even column/.append style={column sep=0.4cm}},
    legend cell align=left,
    nodes near coords,
    nodes near coords style={font=\tiny, rotate=0, anchor=center, yshift=3pt, /pgf/number format/.cd, fixed, fixed zerofill, precision=0},
    tick align=inside
]
\addplot[fill=yellow!55, draw=yellow!50!black] coordinates {
  (Qwen3, 52.60) (InternVL3.5, 18.63) (GPT-5.4, 89.52) (Mean, 46.39)
};
\addlegendentry{\textit{2-reverse}}
\addplot[fill=orange!55, draw=orange!75!black] coordinates {
  (Qwen3, 30.28) (InternVL3.5, 17.36) (GPT-5.4, 75.73) (Mean, 34.20)
};
\addlegendentry{\textit{3-reverse}}
\addplot[fill=biashlcolor!45, draw=biashlcolor!80!black] coordinates {
  (Qwen3, 36.12) (InternVL3.5, 9.49) (GPT-5.4, 76.40) (Mean, 33.53)
};
\addlegendentry{\textit{3-mixed}}
\end{axis}
\end{tikzpicture}
\caption{\textbf{Adding a prior-conflicting (\textit{3-reverse}) or prior-conforming (\textit{3-mixed}) auxiliary edge to \textit{2-reverse} reduces relation-direction accuracy in every family; the \gptfam family stays much higher overall but still loses ${\sim}14$ points from \textit{2-reverse} to \textit{3-reverse}.} Per-family and overall mean $\mathrm{RelAcc}$ (\%) on \textit{2-reverse} (yellow), \textit{3-reverse} (orange), and \textit{3-mixed} (red), averaged over 4 relations and 3 image scales; each family bar is the simple mean over its members (4 for \qwenfam and \internfam, 2 for \gptfam), and ``Mean'' is the simple mean over all 10 models.}
\label{fig:finding1_three_class}
\end{figure}
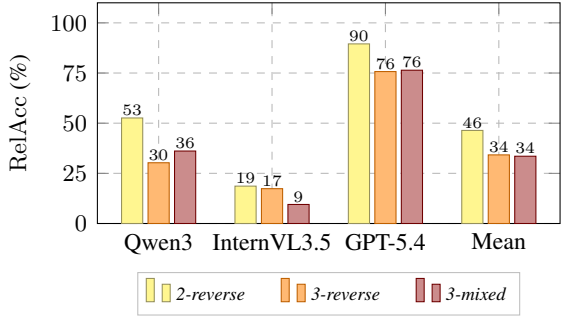

\paragraph{Adding more classes to the prior-conflicting scenario further reduces model performance.}
\figref{fig:finding1_three_class} shows that augmenting the two-class prior-conflicting diagram with a third class further reduces accuracy. The 10-model mean $\mathrm{RelAcc}$ falls from 46.39\% on \textit{2-reverse} to \biashl{$34.20\%$} on \textit{3-reverse} ($-12.19$ \%) and to \biashl{$33.53\%$} on \textit{3-mixed} ($-12.86$ \%). The two relation-direction accuracies of three-class conditions are almost indistinguishable, even though \textit{3-mixed} adds an edge whose direction matches common semantic expectations and could in principle anchor the model toward a more visual interpretation. The family-level pattern, however, is asymmetric in the open-source families: \qwenfam scores higher on \textit{3-mixed} (36.12\%) than on \textit{3-reverse} (30.28\%), while \internfam shows the opposite (\textit{3-mixed}'s 9.49\% vs.\ \textit{3-reverse}'s 17.36\%); the two open-source families' patterns cancel out at the open-source 8-model mean.


\paragraph{Closed-source frontier models substantially reduce the conflict gap while the advantage weakens on harder structural variants.}
We report \gptlarge and \gptmini separately because their behavior departs sharply from the open-source pattern. \gptmini reduces the conflict gap to \biashl{$\Delta = 10.00$ \%} ($\mathrm{RelAcc}_\textit{prior-conform} = 97.95\%$, $\mathrm{RelAcc}_\textit{prior-conflict} = 87.95\%$), about a third of the open-source 8-model mean (33.48\%). \gptlarge nearly closes the gap ($\mathrm{RelAcc}_\textit{prior-conform} = 90.73\%$, $\mathrm{RelAcc}_\textit{prior-conflict} = 91.10\%$; \ceilhl{$\Delta \approx 0$}): unlike every open-source model, it does not lose accuracy when the arrow is reversed. The failure persists, however, on the harder three-class conditions: \gptmini drops to $65.41\%$ on \textit{3-reverse} ($-32.55$ from \textit{prior-conform}), and even \gptlarge drops to $86.05\%$ ($-4.68$). Frontier capacity, therefore, largely overcomes the prior-over-vision failure on the clean two-class contrast, but the bias re-emerges once auxiliary structural context is added.

\subsection{\qtwo: Does the prior-over-vision effect persist across relations and image scales?}
\label{sec:finding2}

\subsubsection{Motivation}
\qone showed that the prior-override effect is large in aggregate. \qtwo asks whether it is uniform along the two axes built into our benchmark: (1) the relation type, and (2) the image size of the diagram. Specifically, if larger images shrank the gap, the failure could be re-attributed to a perceptual bottleneck rather than a post-reading bias. We rule out both.

\subsubsection{Experiments}
\begin{figure}[t]
\centering
\begin{tikzpicture}
\begin{groupplot}[
    group style={group size=1 by 2, vertical sep=0.85cm, x descriptions at=edge bottom},
    width=\columnwidth,
    height=3.5cm,
    ymin=0, ymax=115,
    ytick={0,25,50,75,100},
    symbolic x coords={Aggreg., Compos., Depend., Inherit., Mean},
    xtick=data,
    label style={font=\scriptsize},
    tick label style={font=\tiny},
    grid=major,
    grid style={dashed, gray!55},
    legend cell align=left,
    tick align=inside
]
\nextgroupplot[
    title={\scriptsize Open-source (8-model)},
    ylabel={RelAcc / $\Delta$ (\%)},
    ybar,
    bar width=0.2cm,
    enlarge x limits=0.13,
    tick align=inside
]
\addplot[fill=blue!45, draw=blue!80!black] coordinates {
  (Aggreg., 85.42) (Compos., 77.97) (Depend., 39.14) (Inherit., 73.84) (Mean, 69.09)
};
\addplot[fill=orange!55, draw=orange!75!black] coordinates {
  (Aggreg., 34.71) (Compos., 53.57) (Depend., 30.90) (Inherit., 23.27) (Mean, 35.61)
};
\addplot[fill=biashlcolor!45, draw=biashlcolor!80!black] coordinates {
  (Aggreg., 50.71) (Compos., 24.40) (Depend., 8.24) (Inherit., 50.57) (Mean, 33.48)
};
\nextgroupplot[
    title={\scriptsize \gptfam (2-model)},
    ylabel={RelAcc / $\Delta$ (\%)},
    ybar,
    bar width=0.22cm,
    enlarge x limits=0.13,
    legend style={font=\tiny, draw=gray!50, fill=white, at={(0.5,-0.45)}, anchor=north, legend columns=3, /tikz/every even column/.append style={column sep=0.4cm}},
    tick align=inside
]
\addplot[fill=blue!45, draw=blue!80!black] coordinates {
  (Aggreg., 93.71) (Compos., 86.68) (Depend., 97.39) (Inherit., 99.58) (Mean, 94.34)
};
\addlegendentry{\textit{prior-conform}}
\addplot[fill=orange!55, draw=orange!75!black] coordinates {
  (Aggreg., 87.58) (Compos., 81.93) (Depend., 95.26) (Inherit., 93.34) (Mean, 89.52)
};
\addlegendentry{\textit{prior-conflict}}
\addplot[fill=biashlcolor!45, draw=biashlcolor!80!black] coordinates {
  (Aggreg., 6.13) (Compos., 4.75) (Depend., 2.13) (Inherit., 6.25) (Mean, 4.82)
};
\addlegendentry{$\Delta$}
\end{groupplot}
\end{tikzpicture}
\caption{\textbf{The conflict gap is large and positive for all four UML relations in the open-source 8-model mean (top), with Aggregation and Inheritance largest; the \gptfam family (bottom) keeps the gap below 7\% on every relation.} Per-relation and overall mean $\mathrm{RelAcc}$ (\%) on \textit{prior-conform} (blue) and \textit{prior-conflict} (orange), with the conflict gap $\Delta$ (red); ``Mean'' on each panel is the simple mean over the four relations.}
\label{fig:finding2_relations}
\end{figure}
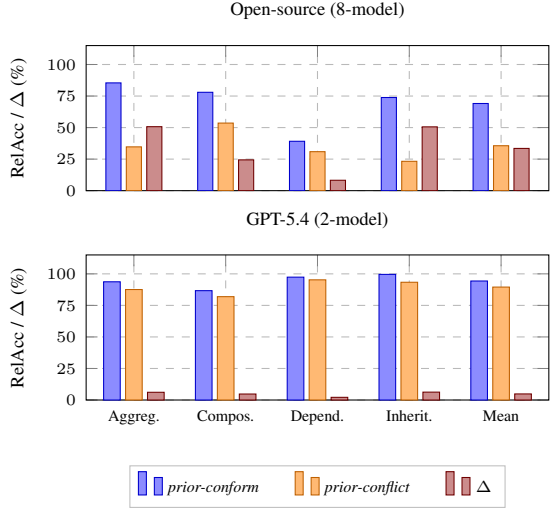

For each of the four relation types (Aggregation, Composition, Dependency, Inheritance), we report the 8-model mean of $\mathrm{RelAcc}$ on \textit{prior-conform} and \textit{prior-conflict} images, averaged across three image scales, with the per-relation conflict gap $\Delta$ (\figref{fig:finding2_relations}). Separately, for each image scale ($1\times$, $1.5\times$, $2\times$), we report the same quantities averaged across all eight models and four relations, and compare $\Delta$ across scales (\figref{fig:finding2_scales}).

\subsubsection{Results}

\begin{figure}[t]
\centering
\begin{tikzpicture}
\begin{groupplot}[
    group style={group size=1 by 2, vertical sep=0.85cm, x descriptions at=edge bottom},
    width=\columnwidth,
    height=3.5cm,
    ymin=0, ymax=115,
    ytick={0,25,50,75,100},
    symbolic x coords={1x, 1.5x, 2x, Mean},
    xtick=data,
    xticklabels={$1\times$, $1.5\times$, $2\times$, Mean},
    label style={font=\scriptsize},
    tick label style={font=\tiny},
    grid=major,
    grid style={dashed, gray!55},
    legend cell align=left,
    tick align=inside
]
\nextgroupplot[
    title={\scriptsize Open-source (8-model)},
    ylabel={RelAcc / $\Delta$ (\%)},
    ybar,
    bar width=0.30cm,
    enlarge x limits=0.20,
    tick align=inside
]
\addplot[fill=blue!45, draw=blue!80!black] coordinates {
  (1x, 68.32) (1.5x, 68.97) (2x, 69.99) (Mean, 69.09)
};
\addplot[fill=orange!55, draw=orange!75!black] coordinates {
  (1x, 35.33) (1.5x, 34.86) (2x, 36.65) (Mean, 35.61)
};
\addplot[fill=biashlcolor!45, draw=biashlcolor!80!black] coordinates {
  (1x, 32.98) (1.5x, 34.11) (2x, 33.34) (Mean, 33.48)
};
\nextgroupplot[
    title={\scriptsize \gptfam (2-model)},
    ylabel={RelAcc / $\Delta$ (\%)},
    ybar,
    bar width=0.30cm,
    enlarge x limits=0.20,
    legend style={font=\tiny, draw=gray!50, fill=white, at={(0.5,-0.45)}, anchor=north, legend columns=3, /tikz/every even column/.append style={column sep=0.4cm}},
    tick align=inside
]
\addplot[fill=blue!45, draw=blue!80!black] coordinates {
  (1x, 96.93) (1.5x, 91.71) (2x, 94.39) (Mean, 94.34)
};
\addlegendentry{\textit{prior-conform}}
\addplot[fill=orange!55, draw=orange!75!black] coordinates {
  (1x, 85.23) (1.5x, 91.02) (2x, 92.32) (Mean, 89.52)
};
\addlegendentry{\textit{prior-conflict}}
\addplot[fill=biashlcolor!45, draw=biashlcolor!80!black] coordinates {
  (1x, 11.69) (1.5x, 0.69) (2x, 2.07) (Mean, 4.82)
};
\addlegendentry{$\Delta$}
\end{groupplot}
\end{tikzpicture}
\caption{\textbf{Quadrupling the image area does not narrow the conflict gap in either family group; the prior-override effect is not a perceptual bottleneck. The open-source 8-model gap stays near 33\% at every scale (top), while the \gptfam family stays under 12\% (bottom).} Per-scale and overall mean $\mathrm{RelAcc}$ (\%) on \textit{prior-conform} (blue) and \textit{prior-conflict} (orange), with the conflict gap $\Delta$ (red); ``Mean'' on each panel is the simple mean over the three scales.}
\label{fig:finding2_scales}
\end{figure}
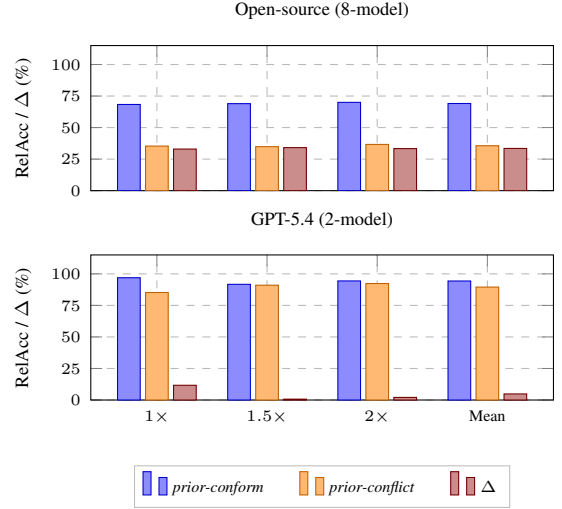

\paragraph{Relations with stronger knowledge priors produce larger conflict gaps.}
\figref{fig:finding2_relations} (top) shows that the open-source conflict gap appears in all four relations, scaling with the strength of knowledge prior. The two largest gaps are on the relations whose class names carry the strongest priors: \textbf{Aggregation} ($\mathrm{RelAcc}_{\textit{prior-conform}}=85.42\%$, \biashl{$\Delta = 50.71\%$}) and \textbf{Inheritance} ($\mathrm{RelAcc}_{\textit{prior-conform}}=73.84\%$, \biashl{$\Delta = 50.57$}). Composition has a smaller gap ($\Delta = 24.40\%$), and Dependency has the smallest ($\Delta = 8.24\%$). The smaller Dependency gap is partly a floor effect as this is the relation with the lowest accuracy. The smaller Composition gap reflects moderate \textit{prior-conform} accuracy (77.97\%) combined with relatively preserved \textit{prior-conflict} accuracy (53.57\%), suggesting that the composition diamond appears more robustly parsed when the arrow direction is reversed. The pattern is mostly consistent: the stronger the knowledge prior, the lower the accuracy when reversing the direction. The \gptfam family (\figref{fig:finding2_relations} bottom) reproduces the same Aggregation/Inheritance signal but at a much smaller magnitude ($\Delta \leq 6.25\%$ on every relation).

\paragraph{Larger images do not narrow the conflict gap.}
\figref{fig:finding2_scales} (top) shows the open-source conflict gap is essentially the same as we quadruple the image area: $\Delta = 32.98$\% at $1\times$, $34.11$\% at $1.5\times$, and $33.34$\% at $2\times$. Both \textit{prior-conform} and \textit{prior-conflict} accuracies rise modestly with scale ($68.32\% \to 69.99\%$ and $35.33\% \to 36.65\%$, respectively). Thus, the visual content is read marginally better at larger scales, but the gap between them is unaffected. Giving the model more pixels does not reduce its reliance on the knowledge prior, and consequently this failure mode is not a perceptual bottleneck. The \gptfam family (\figref{fig:finding2_scales} bottom) is similarly insensitive to scale on $1.5\times$ and $2\times$ ($\Delta \leq 2.07\%$); the larger $\Delta = 11.69\%$ at $1\times$ is driven by \gptmini's instability on the smallest images, not a generic perceptual issue.


\subsection{\qthree: Does scaling the model size fix the prior-over-vision effect?}
\label{sec:finding3}
\subsubsection{Motivation}
Increasing the model size can improve model performance: with more capacity, the model usually performs better and can learn to weigh the visual evidence more heavily than the knowledge prior. \qthree tests this on \textit{2-reverse} and \textit{3-reverse}, and checks the prerequisite skills (class-name reading and arrow-direction reading without knowledge prior) along the same scaling axis. A genuine scaling fix should improve both performance and prerequisite skills monotonically with parameter count.

\subsubsection{Experiments}
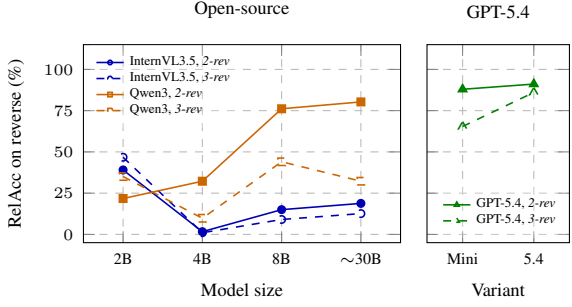
\begin{figure}[!ht]
\centering
\begin{tikzpicture}
\begin{groupplot}[
    group style={group size=2 by 1, horizontal sep=0.35cm, y descriptions at=edge left},
    height=4.2cm,
    ymin=-5, ymax=115,
    ytick={0,25,50,75,100},
    grid=both,
    grid style={dashed, gray!55},
    label style={font=\scriptsize},
    tick label style={font=\tiny},
]
\nextgroupplot[
    title={\scriptsize Open-source},
    width=0.75\columnwidth,
    xmin=0.5, xmax=4.5,
    xtick={1,2,3,4},
    xticklabels={2B, 4B, 8B, {$\sim$30B}},
    xlabel={Model size},
    ylabel={RelAcc on reverse (\%)},
    legend style={font=\fontsize{5}{6}\selectfont, draw=none, fill=white, fill opacity=0.25, text opacity=1, row sep=-3pt, inner sep=2pt, at={(0.02,1.00)}, anchor=north west},
    legend cell align=left,
    legend image post style={scale=0.6},
]
\addplot[mark=*, mark size=1.4pt, semithick, color=blue!70!black] coordinates {(1,39.06) (2,1.66) (3,15.00) (4,18.79)};
\addlegendentry{InternVL3.5, \textit{2-rev}}
\addplot[mark=o, mark size=1.4pt, semithick, dashed, color=blue!70!black] coordinates {(1,46.75) (2,1.02) (3,9.04) (4,12.61)};
\addlegendentry{InternVL3.5, \textit{3-rev}}
\addplot[mark=square*, mark size=1.4pt, semithick, color=orange!80!black] coordinates {(1,21.74) (2,32.23) (3,76.14) (4,80.29)};
\addlegendentry{Qwen3, \textit{2-rev}}
\addplot[mark=square, mark size=1.4pt, semithick, dashed, color=orange!80!black] coordinates {(1,35.08) (2,9.79) (3,44.01) (4,32.24)};
\addlegendentry{Qwen3, \textit{3-rev}}
\nextgroupplot[
    title={\scriptsize \gptfam},
    width=0.45\columnwidth,
    xmin=0.5, xmax=2.5,
    xtick={1,2},
    xticklabels={Mini, 5.4},
    xlabel={Variant},
    legend style={font=\fontsize{5}{6}\selectfont, draw=none, fill=white, fill opacity=0.25, text opacity=1, row sep=-3pt, inner sep=2pt, at={(0.98,0.02)}, anchor=south east},
    legend cell align=left,
    legend image post style={scale=0.6},
]
\addplot[mark=triangle*, mark size=1.8pt, semithick, color=green!50!black] coordinates {(1,87.95) (2,91.10)};
\addlegendentry{\gptfam, \textit{2-rev}}
\addplot[mark=triangle, mark size=1.8pt, semithick, dashed, color=green!50!black] coordinates {(1,65.41) (2,86.05)};
\addlegendentry{\gptfam, \textit{3-rev}}
\end{groupplot}
\end{tikzpicture}
\caption{\textbf{Within-family scaling looks positive for \qwenfam on \textit{2-reverse} but inverse for \internfam (left); on \textit{3-reverse} most larger open-source variants collapse below the 2B.
The \gptfam family (right) sits well above the open-source variants on both conditions, and scales positively from \gptmini to \gptlarge on \textit{3-reverse} ($65.41 \to 86.05$).} Per-variant $\mathrm{RelAcc}$ (\%) on \textit{2-reverse} (solid) and \textit{3-reverse} (dashed) vs.\ within-family scale; shared y-axis across panels.}
\label{fig:finding3_scaling}
\end{figure}

We compare same-family variants, 2B, 4B, 8B, and 32B (or 38B), within \internfam, \qwenfam and GPT separately. For each variant we report (i) $\mathrm{RelAcc}$ on \textit{2-reverse} and \textit{3-reverse} averaged across four relations and three scales (\figref{fig:finding3_scaling}), and (ii) two prerequisite measurements on \textit{prior-free}: $\mathrm{EM}$ (exact class-name recognition) and $\mathrm{RelAcc_\textit{prior-free}}$ (arrow-direction accuracy with prior-free labels), each averaged across four relations and three scales. We adopt thresholds of $\mathrm{EM} \geq 95\%$ (perception intact) and $\mathrm{RelAcc} \geq 50\%$ (arrow-reading above the always-True baseline) and call models passing both the \emph{prerequisite-passing subset} (\figref{fig:finding3_prereq}).

\subsubsection{Results}
\paragraph{Naive scaling helps Qwen3 on \textit{2-reverse} but appears inverse for InternVL3.5.}
\figref{fig:finding3_scaling} shows starkly different scaling patterns across families on \textit{2-reverse}. \qwenfam scales monotonically: 2B ($21.74\%$) $\to$ 4B ($32.23\%$) $\to$ 8B ($76.14\%$) $\to$ 32B (\ceilhl{$80.29\%$}), a $+58.55$-point gain across the family. \internfam, by contrast, appears to scale \emph{inversely}: 2B ($39.06\%$) is uniquely strong, while 4B ($1.66\%$), 8B ($15.00\%$), and 38B ($18.79\%$) all fall well below the 2B; even the 38B model never recovers within 20\% of the 2B. On \textit{3-reverse}, almost every larger variant collapses in both families. Most strikingly, \qwen{32B} drops \biashl{$48.05$\%} from \textit{2-rev} to \textit{3-rev} ($80.29\% \to 32.24\%$), ending below both \qwen{8B} (44.01\%) and \qwen{2B} (35.08\%). On the harder condition, the strongest variant is 2B for InternVL3.5 but 8B for Qwen3.

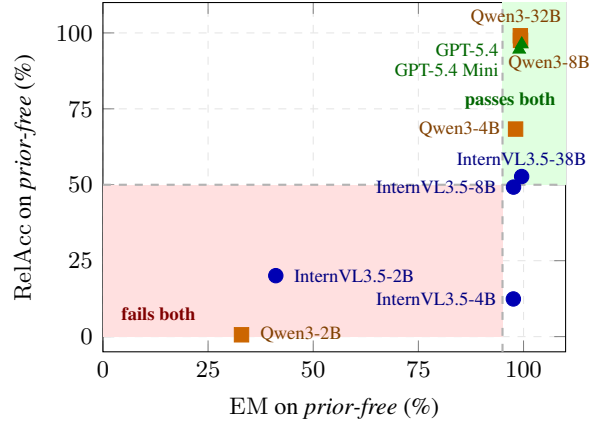
\begin{figure}[t]
\centering
\begin{tikzpicture}
\begin{axis}[
    width=\columnwidth,
    height=6.2cm,
    xmin=0, xmax=110,
    ymin=-5, ymax=110,
    xtick={0,25,50,75,100},
    ytick={0,25,50,75,100},
    xlabel={$\mathrm{EM}$ on \textit{prior-free} (\%)},
    ylabel={$\mathrm{RelAcc}$ on \textit{prior-free} (\%)},
    label style={font=\footnotesize},
    tick label style={font=\footnotesize},
    grid=major,
    grid style={dashed, gray!20},
    clip=false,
]
\fill[green!12] (axis cs:95,50) rectangle (axis cs:110,110);
\fill[red!12]   (axis cs:0,0)   rectangle (axis cs:95,50);
\draw[dashed, gray!60, thick] (axis cs:95,-5)  -- (axis cs:95,110);
\draw[dashed, gray!60, thick] (axis cs:0,50)   -- (axis cs:110,50);
\addplot[only marks, mark=*, mark size=2.7pt, color=blue!70!black] coordinates {
  (41.10, 20.07)
  (97.58, 12.40)
  (97.62, 49.23)
  (99.55, 52.71)
};
\addplot[only marks, mark=square*, mark size=2.7pt, color=orange!80!black] coordinates {
  (32.95,  0.60)
  (98.11, 68.34)
  (99.29, 97.54)
  (99.24, 99.06)
};
\addplot[only marks, mark=triangle*, mark size=2.7pt, color=green!50!black] coordinates {
  (98.97, 94.55)
  (99.58, 96.29)
};
\node[anchor=west,  font=\scriptsize, color=blue!50!black]   at (axis cs:43, 20)    {InternVL3.5-2B};
\node[anchor=east,  font=\scriptsize, color=blue!50!black]   at (axis cs:96, 12.4)  {InternVL3.5-4B};
\node[anchor=east,  font=\scriptsize, color=blue!50!black]   at (axis cs:96, 49.23) {InternVL3.5-8B};
\node[anchor=south, font=\scriptsize, color=blue!50!black]   at (axis cs:99.55, 53) {InternVL3.5-38B};
\node[anchor=west,  font=\scriptsize, color=orange!50!black] at (axis cs:35, 0.6)   {Qwen3-2B};
\node[anchor=east,  font=\scriptsize, color=orange!50!black] at (axis cs:97, 68.34) {Qwen3-4B};
\node[anchor=west,  font=\scriptsize, color=orange!50!black] at (axis cs:94, 90)  {Qwen3-8B};
\node[anchor=west,  font=\scriptsize, color=orange!50!black] at (axis cs:85, 105) {Qwen3-32B};
\node[anchor=east, font=\scriptsize, color=green!40!black] at (axis cs:96.5, 94.55) {\gptlarge};
\node[anchor=east, font=\scriptsize, color=green!40!black] at (axis cs:96.5, 88)    {\gptmini};
\node[anchor=north east, font=\scriptsize\bfseries, color=green!40!black] at (axis cs:110, 84) {passes both};
\node[anchor=south west, font=\scriptsize\bfseries, color=red!50!black]   at (axis cs:2, 2)     {fails both};
\end{axis}
\end{tikzpicture}
\caption{\textbf{Most \internfam variants fail the prerequisite skills required to interpret a low \textit{2-reverse} score as bias; only \intern{38B} and three of four \qwenfam variants pass both, while both closed-source \gptfam models pass comfortably.} Dashed thresholds at $x{=}95$ and $y{=}50$; marker color = family (blue: \internfam; orange: \qwenfam; green: \gptfam).}
\label{fig:finding3_prereq}
\end{figure}

\paragraph{InternVL3.5's apparent inverse scaling reflects skill deficit, not true scaling failure.}
\figref{fig:finding3_prereq} reveals the catch behind \figref{fig:finding3_scaling}: most \internfam variants lack the underlying skills needed to interpret a low \textit{2-reverse} score as bias. Only \intern{38B} passes both prerequisites; \intern{2B} fails both, while \intern{4B} (12.40\%) and \intern{8B} (49.23\%) pass class-recognition but fail arrow-reading without knowledge prior. \qwenfam, by contrast, has three of four variants passing both prerequisites (only \qwen{2B} fails). The apparent inverse scaling within \internfam is therefore largely an artifact: when 4B and 8B cannot read arrows even without a misleading prior, their \textit{2-reverse} accuracy is uninformative about scaling on the bias task. Within the four-model prerequisite-passing subset (\qwen{4B}/8B/32B + \intern{38B}), the headline conflict gap is $\Delta = 32.94$ \%, essentially identical to the all-eight 33.48 \%, so the prior-override claim from \qone is robust.


\section{Conclusion}

We set out to ask whether current VLMs read UML class diagrams or instead answer from pretrained knowledge priors about how named classes typically relate. Across a controlled benchmark evaluated on 8 open-source VLMs from two families (InternVL3.5, Qwen3) and two closed-source frontier models (\gptlarge, \gptmini), we find that reversing a single arrow drops relation accuracy by 33.48\% on average across open-source models, showing that they substitute priors for diagrams in the post-reading inference step. Frontier models reduce this clean two-class gap: \gptmini cuts it to $\Delta = 10.00\%$, and \gptlarge nearly closes it. However, harder three-class variants remain challenging, with average drops of 45.28\% for open-source models and 18.62\% for the \gptlarge family, indicating that even frontier models are not immune once auxiliary structural context is added.

These findings are robust across the two axes built into our benchmark. Quadrupling the image area does not narrow the conflict gap (${\sim}33$\% at every scale), and the gap is largest for relations whose class names carry strong priors (\eg \textit{Aggregation}, \textit{Inheritance}), producing drops of $\Delta \approx 51$\% each. Within open-source families, more parameters do not necessarily improve performance. While Qwen3 shows a steady increase in $\mathrm{RelAcc}_{\textit{conflict}}$ as model size grows, InternVL3.5 drops sharply from 2B to 4B, which can be attributed to weak class-name reading and $\mathrm{RelAcc}_{\textit{prior-free}}$.


\section*{Limitations}
This benchmark is designed for controlled diagnosis of prior-over-visual override in VLMs, and its current scope is intentionally narrow in two respects. First, each accepted diagram is paired with exactly one target relation query over an ordered class pair, so the evaluation measures targeted relation judgment rather than full-diagram parsing or multi-query reasoning. Second, the headline results are reported after aggregating across the four relation types and, for slice-level comparisons, over the models with full protocol coverage. This means the aggregate trends should not be read as implying that all UML relations, model families, or image scales fail in exactly the same way. These choices improve comparability for the main controlled analyses, but some finer-grained relation-specific variation is left to future work.

\bibliography{custom}

\clearpage
\appendix
\section{Additional examples}
\label{sec:appendix_examples_prompts}
\label{sec:appendix_3mixed}
\label{sec:benchmark}
\begin{figure}[!ht]
    \centering
    \includegraphics[width=.90\linewidth]{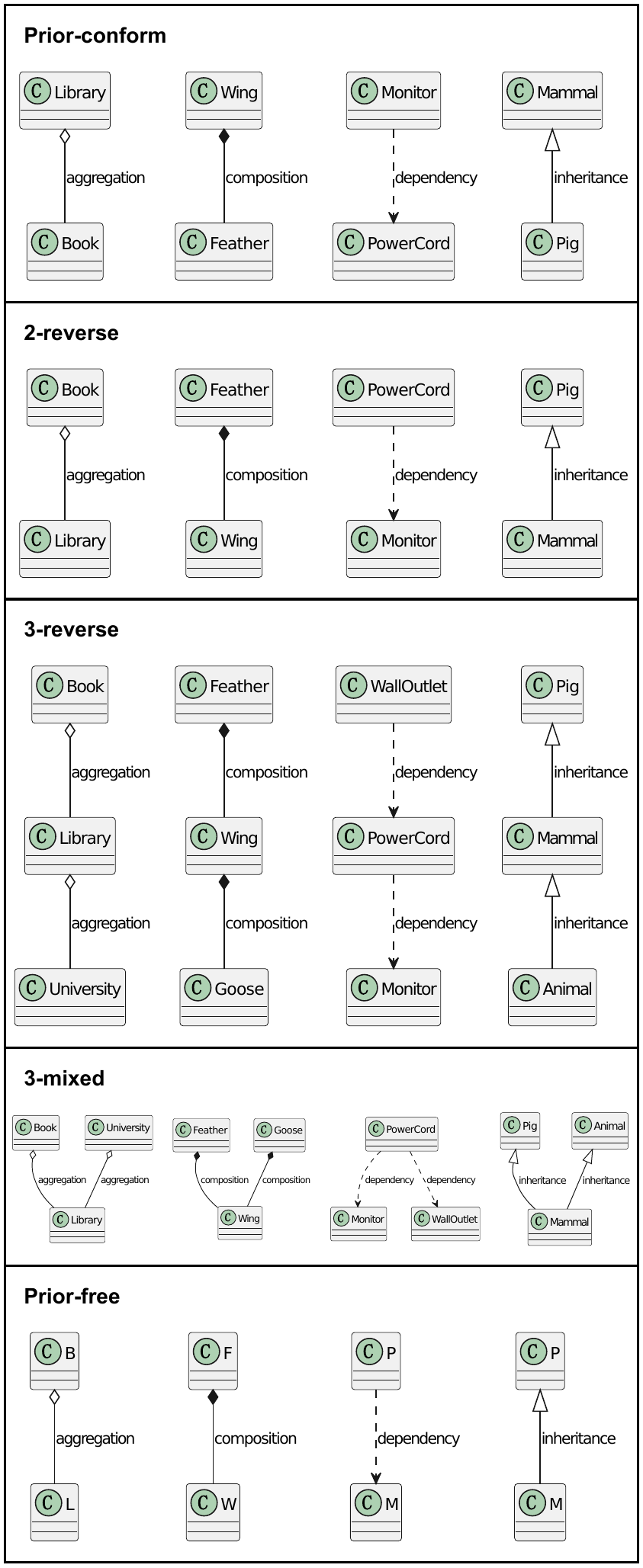}
    \caption{\textbf{Examples for each condition type.} These examples include the additional prior-conflict subsets and the prior-free configurations omitted from the compact main-text figure.}
    \label{fig:full_examples}
\end{figure}

\figref{fig:full_examples} expands the condition illustration in \figref{fig:conditions}. It shows representative diagrams for the prior-conforming, prior-conflicting, and prior-free settings, including the three-class variants used to test whether an added auxiliary relation changes model reliance on the diagram.

\section{Detailed Experimental Setup}
\label{sec:infra}

We evaluate eight open-source VLMs across two families, \texttt{InternVL3.5-\{2B,4B,8B,38B\}} and \texttt{Qwen3-\{2B,4B,8B,32B\}}, plus two closed-source frontier models, \gptlarge and \gptmini. We query each model on all five conditions: prior-conforming, \textit{2-reverse} / \textit{3-reverse} / \textit{3-mixed} (prior-conflicting), and \textit{prior-free}. Each image is evaluated at three image scales ($1\times$, $1.5\times$, $2\times$). We run experiments on 4$\times$ NVIDIA L40S GPUs (48~GB VRAM each), serving open-source models through vLLM and querying closed-source models through their APIs under the same prompt template and output schema.
The open-source inference required approximately 120 GPU-hours, and closed-source API inference cost approximately 200 USD.

\paragraph{Parsing.}
We pass all raw model outputs through a deterministic parser before scoring. The parser handles common formatting variation (\eg punctuation wrappers, sentence envelopes around the answer, paraphrases of True/False, and JSON parsing for class-name recovery) and produces a normalized prediction. Outputs that cannot be resolved unambiguously into $\{$True, False, Unknown$\}$ for the relation query (or into a parsed class-name set for class-name recovery) are conservatively mapped to Unknown.

\section{UML Diagram Rendering}
\label{sec:appendix_rendering}

All diagrams in our benchmark are rendered with PlantUML to ensure a consistent and reproducible visual style~\cite{Roques_PlantUML_Software}. PlantUML is distributed under the GNU General Public License v3.0 (GPL-3.0), and we use it only as a rendering tool for diagrams generated from our own structured specifications. We first represent each accepted UML instance as a structured textual specification, then render it into an image using the same PlantUML pipeline for all conditions. This keeps the diagram style, class-box layout, arrow notation, and relation labels consistent across prior-conforming, prior-conflicting, and \textit{prior-free} examples. For the image-scale analysis, we resize the same rendered diagrams to the evaluated resolutions, so the scale variants differ only in image size and not in UML content.

\section{Benchmark}
\label{sec:appendix_dataset_size}

\tabref{tab:dataset_size} summarizes the number of benchmark instances used in our evaluation. We count one \emph{base instance} as one UML diagram paired with one target relation-direction query before image scaling. Since every base instance is rendered at three image scales ($1\times$, $1.5\times$, and $2\times$), the number of rendered inputs is three times the base-instance count. Overall, the benchmark contains 27,156 base instances and 81,468 rendered image-query inputs across the five evaluated conditions.

The \textit{prior-free} condition is smaller than the other conditions because it is constructed as a diagnostic control rather than sampled to match the full semantic benchmark size. We derive it by replacing the original semantic class names with their uppercase-letter abbreviations, such as \texttt{GoldenRetriever} $\rightarrow$ \texttt{GR} and \texttt{Dog} $\rightarrow$ \texttt{D}. This many-to-one transformation removes class-name semantics, but it also causes many originally distinct diagrams to collapse into the same abbreviated class pair. We therefore keep only unique valid abbreviated instances and discard cases where abbreviation creates duplicate class names within the same diagram. The resulting counts reflect the number of distinct \textit{prior-free} diagrams after this deduplication step.

\begin{table}[h]
\centering
\scriptsize
\setlength{\tabcolsep}{3.5pt}
\renewcommand{\arraystretch}{1.05}
\resizebox{\columnwidth}{!}{%
\begin{tabular}{lrrrrrr}
\toprule
\textbf{Condition} & \textbf{Agg.} & \textbf{Comp.} & \textbf{Dep.} & \textbf{Inh.} & \textbf{Base} & \textbf{Rendered} \\
\midrule
\textit{prior-free}    &   338 &   373 & 1,044 & 1,401 &  3,156 &  9,468 \\
\textit{prior-conform} & 1,200 & 1,200 & 1,300 & 2,300 &  6,000 & 18,000 \\
\textit{2-reverse}     & 1,200 & 1,200 & 1,300 & 2,300 &  6,000 & 18,000 \\
\textit{3-reverse}     & 1,200 & 1,200 & 1,300 & 2,300 &  6,000 & 18,000 \\
\textit{3-mixed}       & 1,200 & 1,200 & 1,300 & 2,300 &  6,000 & 18,000 \\
\midrule
\textbf{Total}         & 5,138 & 5,173 & 6,244 & 10,601 & 27,156 & 81,468 \\
\bottomrule
\end{tabular}%
}
\caption{Benchmark size by condition and UML relation. Counts under Agg., Comp., Dep., and Inh. are base instances before image scaling. Rendered counts include the three evaluated image scales ($1\times$, $1.5\times$, and $2\times$).}
\label{tab:dataset_size}
\end{table}

\section{Full per-cell results}
\label{sec:appendix_main_table}
\tabref{tab:main_table} reports \textbf{\textcolor{relaccRed}{RelAcc}} and \textbf{EM} for every (model, condition, relation, image scale) cell across all five conditions in the table, including the \textit{3-mixed} and \textit{3-reverse} conditions shown in \secref{sec:appendix_3mixed}, and adds the closed-source \gptlarge and \gptmini blocks for completeness. The aggregated views in the main text (\figref{fig:finding1_conflict}, \figref{fig:finding1_three_class}, \figref{fig:finding2_relations}, \figref{fig:finding2_scales}, \figref{fig:finding3_scaling}, \figref{fig:finding3_prereq}) are derived by averaging over relations and/or scales as noted in their captions.

\section{Evaluation prompts}
\label{sec:appendix_prompts}
The prompts are deliberately designed to minimize ambiguity in the visual interpretation task. They first make the notation explicit by describing the visible components of each class box and clarifying that the green circled ``C'' is only a class-type icon, not part of the relation semantics. They then define the four UML relations in the benchmark, including inheritance, aggregation, composition, and dependency, and specify how each arrowhead or line ending should be interpreted. Beyond explaining the notation, the prompts repeatedly emphasize that the diagram itself is the only valid source of evidence. The model is instructed to inspect the arrow or line direction carefully, disregard external knowledge about the class names, and avoid making judgments from common software-design assumptions unless the relation is visually supported.

This prompt design makes the evaluation intentionally conservative. The goal is not to test whether models fail because the diagram is underspecified or because the notation is unfamiliar. Instead, the instructions give the model the relevant visual rules in advance and explicitly warn it against relying on semantic priors. Therefore, when a model still predicts the prior-consistent relation under conflict conditions, the error is harder to attribute to prompt ambiguity. It more directly reflects a failure to privilege visual evidence over learned software-design priors. The prompts used in the experiments are shown below, where terms in angle brackets (<>) denote instance-specific placeholders (\figref{fig:prompt_system}, \figref{fig:prompt_stage1}, \figref{fig:prompt_stage2}, \figref{fig:prompt_rules_templates}).

\definecolor{modelRow}{HTML}{E9ECEF}
\definecolor{ablationRow}{HTML}{F7F4D7}
\definecolor{forwardRow}{HTML}{EAF3FF}
\definecolor{reverseRow}{HTML}{FFF1E8}
\definecolor{mixedRow}{HTML}{EEF8EE}
\begin{table*}[h]
  \centering
  \scriptsize
  \setlength{\tabcolsep}{2.4pt}
  \renewcommand{\arraystretch}{1.05}
  \resizebox{\textwidth}{!}{%
  \begin{tabular}{>{\columncolor{white}}c l @{\hspace{6pt}} cc @{\hspace{2pt}} cc @{\hspace{2pt}} cc @{\hspace{8pt}} cc @{\hspace{2pt}} cc @{\hspace{2pt}} cc @{\hspace{8pt}} cc @{\hspace{2pt}} cc @{\hspace{2pt}} cc @{\hspace{8pt}} cc @{\hspace{2pt}} cc @{\hspace{2pt}} cc}
    \toprule
    \textbf{Size} & \textbf{Setting} & \multicolumn{6}{c}{\textbf{Aggregation}} & \multicolumn{6}{c}{\textbf{Composition}} & \multicolumn{6}{c}{\textbf{Dependency}} & \multicolumn{6}{c}{\textbf{Inheritance}} \\
    \rowcolor{black!2}
    {} & {} & \multicolumn{2}{c}{$1\times$} & \multicolumn{2}{c}{$1.5\times$} & \multicolumn{2}{c}{$2\times$} & \multicolumn{2}{c}{$1\times$} & \multicolumn{2}{c}{$1.5\times$} & \multicolumn{2}{c}{$2\times$} & \multicolumn{2}{c}{$1\times$} & \multicolumn{2}{c}{$1.5\times$} & \multicolumn{2}{c}{$2\times$} & \multicolumn{2}{c}{$1\times$} & \multicolumn{2}{c}{$1.5\times$} & \multicolumn{2}{c}{$2\times$} \\
    \rowcolor{black!4}
    {} & {} & \relaccval{RelAcc} & EM & \relaccval{RelAcc} & EM & \relaccval{RelAcc} & EM & \relaccval{RelAcc} & EM & \relaccval{RelAcc} & EM & \relaccval{RelAcc} & EM & \relaccval{RelAcc} & EM & \relaccval{RelAcc} & EM & \relaccval{RelAcc} & EM & \relaccval{RelAcc} & EM & \relaccval{RelAcc} & EM & \relaccval{RelAcc} & EM \\
    \rowcolor{modelRow}
    \multicolumn{26}{c}{\textbf{InternVL3.5}} \\
    \cmidrule{1-26}
    \rowcolor{ablationRow}
    {} & prior-free & \relaccval{21.30} & 25.44 & \relaccval{20.12} & 23.67 & \relaccval{17.75} & 22.19 & \relaccval{12.06} & 16.89 & \relaccval{9.65} & 15.01 & \relaccval{9.92} & 13.94 & \relaccval{17.34} & 67.05 & \relaccval{17.24} & 65.23 & \relaccval{17.62} & 65.52 & \relaccval{34.19} & 61.74 & \relaccval{32.12} & 58.82 & \relaccval{31.55} & 57.74 \\
    \rowcolor{forwardRow}
    {} & prior-conform & \relaccval{96.17} & 99.50 & \relaccval{96.25} & 99.50 & \relaccval{96.00} & 99.33 & \relaccval{90.83} & 99.08 & \relaccval{90.92} & 99.00 & \relaccval{91.00} & 99.17 & \relaccval{40.08} & 95.62 & \relaccval{39.85} & 95.00 & \relaccval{39.23} & 95.15 & \relaccval{48.83} & 96.43 & \relaccval{52.74} & 96.39 & \relaccval{52.70} & 96.39 \\
    \rowcolor{reverseRow}
    {} & 2-reverse & \relaccval{44.00} & 98.50 & \relaccval{45.08} & 98.50 & \relaccval{46.08} & 98.67 & \relaccval{71.67} & 99.42 & \relaccval{71.67} & 99.42 & \relaccval{73.33} & 99.33 & \relaccval{27.85} & 96.15 & \relaccval{27.54} & 96.46 & \relaccval{28.15} & 96.08 & \relaccval{10.09} & 96.61 & \relaccval{11.61} & 96.65 & \relaccval{11.65} & 96.74 \\
    \rowcolor{reverseRow}
    {} & 3-reverse & \relaccval{50.33} & 98.00 & \relaccval{51.08} & 98.25 & \relaccval{51.08} & 98.00 & \relaccval{67.67} & 99.08 & \relaccval{67.08} & 98.83 & \relaccval{66.92} & 98.83 & \relaccval{34.62} & 95.54 & \relaccval{35.08} & 95.54 & \relaccval{34.08} & 95.54 & \relaccval{35.13} & 91.65 & \relaccval{34.43} & 90.70 & \relaccval{33.48} & 90.35 \\
    \rowcolor{mixedRow}
    \multirow{-5}{*}{2B} & 3-mixed & \relaccval{5.92} & 97.92 & \relaccval{6.50} & 97.92 & \relaccval{6.42} & 97.92 & \relaccval{19.92} & 98.67 & \relaccval{20.83} & 98.75 & \relaccval{21.67} & 98.67 & \relaccval{32.77} & 97.46 & \relaccval{30.46} & 97.54 & \relaccval{31.85} & 97.62 & \relaccval{58.26} & 96.43 & \relaccval{55.65} & 96.65 & \relaccval{54.96} & 96.43 \\
    \cmidrule{1-26}
    \rowcolor{ablationRow}
    {} & prior-free & \relaccval{36.69} & 96.45 & \relaccval{36.69} & 96.75 & \relaccval{36.09} & 96.15 & \relaccval{12.87} & 98.66 & \relaccval{10.99} & 99.46 & \relaccval{11.26} & 99.46 & \relaccval{0.00} & 97.61 & \relaccval{0.00} & 97.51 & \relaccval{0.00} & 97.61 & \relaccval{1.07} & 96.86 & \relaccval{1.50} & 97.14 & \relaccval{1.57} & 97.29 \\
    \rowcolor{forwardRow}
    {} & prior-conform & \relaccval{56.00} & 99.25 & \relaccval{59.58} & 99.17 & \relaccval{59.67} & 99.33 & \relaccval{29.50} & 99.67 & \relaccval{32.08} & 99.42 & \relaccval{31.17} & 99.42 & \relaccval{0.69} & 97.46 & \relaccval{0.77} & 97.38 & \relaccval{0.54} & 97.15 & \relaccval{57.09} & 97.96 & \relaccval{57.17} & 98.04 & \relaccval{57.39} & 98.13 \\
    \rowcolor{reverseRow}
    {} & 2-reverse & \relaccval{1.17} & 99.17 & \relaccval{1.17} & 99.25 & \relaccval{1.17} & 99.17 & \relaccval{4.92} & 99.33 & \relaccval{5.75} & 99.33 & \relaccval{5.58} & 99.33 & \relaccval{0.08} & 97.31 & \relaccval{0.00} & 97.38 & \relaccval{0.08} & 97.08 & \relaccval{0.00} & 98.00 & \relaccval{0.00} & 97.78 & \relaccval{0.00} & 97.70 \\
    \rowcolor{reverseRow}
    {} & 3-reverse & \relaccval{0.25} & 99.08 & \relaccval{0.42} & 99.17 & \relaccval{0.42} & 99.08 & \relaccval{3.83} & 99.17 & \relaccval{3.75} & 99.17 & \relaccval{3.58} & 99.17 & \relaccval{0.00} & 97.92 & \relaccval{0.00} & 98.00 & \relaccval{0.00} & 98.00 & \relaccval{0.00} & 97.43 & \relaccval{0.00} & 97.48 & \relaccval{0.00} & 97.43 \\
    \rowcolor{mixedRow}
    \multirow{-5}{*}{4B} & 3-mixed & \relaccval{0.00} & 98.83 & \relaccval{0.00} & 98.75 & \relaccval{0.00} & 98.92 & \relaccval{0.17} & 98.92 & \relaccval{0.08} & 99.00 & \relaccval{0.17} & 98.83 & \relaccval{0.00} & 97.69 & \relaccval{0.00} & 97.77 & \relaccval{0.00} & 97.92 & \relaccval{0.00} & 97.04 & \relaccval{0.00} & 97.09 & \relaccval{0.00} & 97.04 \\
    \cmidrule{1-26}
    \rowcolor{ablationRow}
    {} & prior-free & \relaccval{56.21} & 98.52 & \relaccval{54.14} & 98.52 & \relaccval{55.03} & 98.52 & \relaccval{66.49} & 98.93 & \relaccval{60.86} & 99.20 & \relaccval{59.52} & 98.93 & \relaccval{0.19} & 95.11 & \relaccval{0.10} & 95.21 & \relaccval{0.10} & 95.02 & \relaccval{81.51} & 97.72 & \relaccval{77.59} & 97.93 & \relaccval{78.94} & 97.79 \\
    \rowcolor{forwardRow}
    {} & prior-conform & \relaccval{87.75} & 99.83 & \relaccval{88.58} & 99.83 & \relaccval{88.58} & 99.83 & \relaccval{86.25} & 99.83 & \relaccval{85.58} & 99.67 & \relaccval{85.08} & 99.67 & \relaccval{1.00} & 98.08 & \relaccval{1.15} & 98.08 & \relaccval{1.08} & 98.08 & \relaccval{96.00} & 98.65 & \relaccval{95.96} & 98.61 & \relaccval{95.91} & 98.65 \\
    \rowcolor{reverseRow}
    {} & 2-reverse & \relaccval{17.17} & 99.67 & \relaccval{15.75} & 99.75 & \relaccval{16.33} & 99.75 & \relaccval{35.75} & 99.83 & \relaccval{34.67} & 99.75 & \relaccval{34.42} & 99.83 & \relaccval{0.00} & 98.77 & \relaccval{0.00} & 98.92 & \relaccval{0.00} & 98.92 & \relaccval{9.35} & 98.70 & \relaccval{8.00} & 98.65 & \relaccval{8.57} & 98.61 \\
    \rowcolor{reverseRow}
    {} & 3-reverse & \relaccval{9.17} & 99.67 & \relaccval{8.50} & 99.67 & \relaccval{8.75} & 99.67 & \relaccval{26.08} & 99.58 & \relaccval{25.67} & 99.67 & \relaccval{26.50} & 99.58 & \relaccval{0.23} & 98.08 & \relaccval{0.38} & 98.31 & \relaccval{0.38} & 98.08 & \relaccval{1.13} & 98.52 & \relaccval{0.74} & 98.43 & \relaccval{0.96} & 98.43 \\
    \rowcolor{mixedRow}
    \multirow{-5}{*}{8B} & 3-mixed & \relaccval{0.00} & 99.50 & \relaccval{0.00} & 99.50 & \relaccval{0.00} & 99.50 & \relaccval{0.33} & 99.17 & \relaccval{0.25} & 99.17 & \relaccval{0.25} & 99.25 & \relaccval{0.23} & 98.23 & \relaccval{0.31} & 98.38 & \relaccval{0.31} & 98.23 & \relaccval{0.43} & 98.61 & \relaccval{0.48} & 98.57 & \relaccval{0.43} & 98.57 \\
    \cmidrule{1-26}
    \rowcolor{ablationRow}
    {} & prior-free & \relaccval{99.41} & 100.00 & \relaccval{99.11} & 100.00 & \relaccval{99.11} & 100.00 & \relaccval{96.78} & 100.00 & \relaccval{94.64} & 100.00 & \relaccval{95.17} & 100.00 & \relaccval{9.00} & 99.23 & \relaccval{10.25} & 99.14 & \relaccval{10.82} & 99.14 & \relaccval{4.07} & 99.00 & \relaccval{7.14} & 99.07 & \relaccval{7.00} & 99.00 \\
    \rowcolor{forwardRow}
    {} & prior-conform & \relaccval{98.42} & 99.83 & \relaccval{98.17} & 99.83 & \relaccval{98.33} & 99.83 & \relaccval{95.67} & 99.75 & \relaccval{95.83} & 99.75 & \relaccval{95.75} & 99.75 & \relaccval{18.46} & 98.00 & \relaccval{17.92} & 98.23 & \relaccval{17.85} & 98.15 & \relaccval{35.52} & 99.17 & \relaccval{39.87} & 99.22 & \relaccval{38.13} & 99.17 \\
    \rowcolor{reverseRow}
    {} & 2-reverse & \relaccval{24.50} & 100.00 & \relaccval{22.08} & 100.00 & \relaccval{21.83} & 100.00 & \relaccval{52.08} & 99.50 & \relaccval{51.83} & 99.50 & \relaccval{51.17} & 99.50 & \relaccval{0.54} & 98.38 & \relaccval{0.77} & 98.31 & \relaccval{0.62} & 98.31 & \relaccval{0.04} & 98.83 & \relaccval{0.04} & 98.87 & \relaccval{0.00} & 98.87 \\
    \rowcolor{reverseRow}
    {} & 3-reverse & \relaccval{20.17} & 99.75 & \relaccval{17.67} & 99.75 & \relaccval{18.25} & 99.75 & \relaccval{30.08} & 99.50 & \relaccval{27.17} & 99.42 & \relaccval{26.33} & 99.58 & \relaccval{3.38} & 99.00 & \relaccval{3.46} & 99.00 & \relaccval{3.69} & 99.00 & \relaccval{0.39} & 98.61 & \relaccval{0.35} & 98.57 & \relaccval{0.35} & 98.57 \\
    \rowcolor{mixedRow}
    \multirow{-5}{*}{38B} & 3-mixed & \relaccval{3.08} & 99.83 & \relaccval{2.83} & 99.83 & \relaccval{2.75} & 99.83 & \relaccval{18.33} & 99.17 & \relaccval{18.67} & 99.17 & \relaccval{18.67} & 99.17 & \relaccval{11.92} & 98.54 & \relaccval{12.00} & 98.15 & \relaccval{12.08} & 98.31 & \relaccval{2.22} & 98.70 & \relaccval{2.13} & 98.83 & \relaccval{2.22} & 98.70 \\
    \midrule
    \rowcolor{modelRow}
    \multicolumn{26}{c}{\textbf{Qwen3}} \\
    \cmidrule{1-26}
    \rowcolor{ablationRow}
    {} & prior-free & \relaccval{0.30} & 25.74 & \relaccval{0.30} & 19.53 & \relaccval{0.30} & 17.75 & \relaccval{0.27} & 20.91 & \relaccval{0.27} & 18.23 & \relaccval{0.80} & 14.48 & \relaccval{1.34} & 58.24 & \relaccval{1.44} & 56.13 & \relaccval{1.25} & 57.85 & \relaccval{0.36} & 36.55 & \relaccval{0.21} & 32.48 & \relaccval{0.29} & 37.54 \\
    \rowcolor{forwardRow}
    {} & prior-conform & \relaccval{42.00} & 99.83 & \relaccval{46.33} & 99.83 & \relaccval{54.50} & 99.58 & \relaccval{23.08} & 99.58 & \relaccval{27.33} & 99.58 & \relaccval{32.08} & 99.58 & \relaccval{27.23} & 98.08 & \relaccval{27.92} & 98.38 & \relaccval{34.23} & 98.23 & \relaccval{49.61} & 98.22 & \relaccval{51.09} & 98.09 & \relaccval{58.09} & 98.70 \\
    \rowcolor{reverseRow}
    {} & 2-reverse & \relaccval{17.00} & 99.58 & \relaccval{21.58} & 99.50 & \relaccval{29.83} & 99.33 & \relaccval{10.08} & 99.75 & \relaccval{13.33} & 99.67 & \relaccval{15.42} & 99.50 & \relaccval{34.00} & 99.23 & \relaccval{34.23} & 98.77 & \relaccval{38.23} & 99.38 & \relaccval{14.48} & 98.52 & \relaccval{13.78} & 98.26 & \relaccval{18.91} & 98.39 \\
    \rowcolor{reverseRow}
    {} & 3-reverse & \relaccval{33.50} & 99.67 & \relaccval{39.08} & 99.67 & \relaccval{45.08} & 99.42 & \relaccval{10.50} & 99.50 & \relaccval{14.50} & 99.42 & \relaccval{23.75} & 99.50 & \relaccval{63.31} & 98.15 & \relaccval{57.69} & 98.38 & \relaccval{65.23} & 98.85 & \relaccval{16.96} & 97.52 & \relaccval{24.70} & 97.61 & \relaccval{26.61} & 98.26 \\
    \rowcolor{mixedRow}
    \multirow{-5}{*}{2B} & 3-mixed & \relaccval{3.83} & 99.33 & \relaccval{3.75} & 99.50 & \relaccval{14.50} & 99.33 & \relaccval{2.75} & 99.00 & \relaccval{3.42} & 99.50 & \relaccval{4.58} & 99.33 & \relaccval{42.08} & 99.15 & \relaccval{55.15} & 98.92 & \relaccval{61.46} & 99.08 & \relaccval{36.43} & 97.78 & \relaccval{27.48} & 98.13 & \relaccval{43.22} & 97.48 \\
    \cmidrule{1-26}
    \rowcolor{ablationRow}
    {} & prior-free & \relaccval{86.39} & 98.82 & \relaccval{84.32} & 97.93 & \relaccval{83.43} & 97.04 & \relaccval{84.45} & 99.20 & \relaccval{84.72} & 99.46 & \relaccval{90.08} & 99.46 & \relaccval{17.72} & 97.61 & \relaccval{16.57} & 97.80 & \relaccval{14.85} & 97.51 & \relaccval{86.72} & 97.79 & \relaccval{83.65} & 97.64 & \relaccval{87.22} & 97.07 \\
    \rowcolor{forwardRow}
    {} & prior-conform & \relaccval{95.75} & 99.75 & \relaccval{95.58} & 99.75 & \relaccval{94.92} & 99.75 & \relaccval{94.83} & 99.50 & \relaccval{93.92} & 99.58 & \relaccval{95.50} & 99.58 & \relaccval{25.31} & 98.46 & \relaccval{23.85} & 98.38 & \relaccval{28.85} & 97.69 & \relaccval{96.26} & 98.57 & \relaccval{95.26} & 98.70 & \relaccval{97.57} & 98.57 \\
    \rowcolor{reverseRow}
    {} & 2-reverse & \relaccval{23.33} & 99.92 & \relaccval{21.17} & 99.83 & \relaccval{23.33} & 99.75 & \relaccval{71.75} & 99.92 & \relaccval{68.75} & 99.83 & \relaccval{73.83} & 99.92 & \relaccval{6.62} & 99.15 & \relaccval{5.54} & 99.31 & \relaccval{10.31} & 99.31 & \relaccval{25.70} & 98.43 & \relaccval{22.57} & 98.52 & \relaccval{33.87} & 98.87 \\
    \rowcolor{reverseRow}
    {} & 3-reverse & \relaccval{2.08} & 99.42 & \relaccval{3.83} & 99.67 & \relaccval{4.58} & 100.00 & \relaccval{28.75} & 99.67 & \relaccval{30.58} & 99.92 & \relaccval{44.75} & 99.92 & \relaccval{0.92} & 97.15 & \relaccval{0.77} & 97.54 & \relaccval{0.69} & 97.62 & \relaccval{0.39} & 98.78 & \relaccval{0.04} & 99.09 & \relaccval{0.04} & 99.00 \\
    \rowcolor{mixedRow}
    \multirow{-5}{*}{4B} & 3-mixed & \relaccval{6.50} & 99.50 & \relaccval{4.25} & 99.67 & \relaccval{3.83} & 99.67 & \relaccval{41.67} & 99.50 & \relaccval{30.67} & 99.42 & \relaccval{31.17} & 99.50 & \relaccval{3.08} & 98.23 & \relaccval{5.08} & 98.85 & \relaccval{12.31} & 98.92 & \relaccval{1.13} & 98.30 & \relaccval{0.04} & 98.39 & \relaccval{1.48} & 98.17 \\
    \cmidrule{1-26}
    \rowcolor{ablationRow}
    {} & prior-free & \relaccval{99.11} & 99.11 & \relaccval{98.82} & 99.11 & \relaccval{97.93} & 99.11 & \relaccval{96.25} & 99.46 & \relaccval{95.44} & 99.46 & \relaccval{94.37} & 99.46 & \relaccval{96.65} & 98.85 & \relaccval{97.13} & 99.33 & \relaccval{98.08} & 99.52 & \relaccval{99.00} & 99.21 & \relaccval{98.79} & 99.21 & \relaccval{98.86} & 99.64 \\
    \rowcolor{forwardRow}
    {} & prior-conform & \relaccval{99.25} & 99.92 & \relaccval{99.33} & 99.92 & \relaccval{99.58} & 100.00 & \relaccval{98.58} & 99.83 & \relaccval{98.42} & 99.75 & \relaccval{98.92} & 99.42 & \relaccval{98.38} & 99.62 & \relaccval{98.31} & 99.54 & \relaccval{99.08} & 99.62 & \relaccval{99.52} & 99.52 & \relaccval{99.52} & 99.52 & \relaccval{99.39} & 99.39 \\
    \rowcolor{reverseRow}
    {} & 2-reverse & \relaccval{63.17} & 99.67 & \relaccval{62.33} & 99.58 & \relaccval{64.83} & 99.75 & \relaccval{83.67} & 100.00 & \relaccval{82.42} & 100.00 & \relaccval{85.08} & 99.50 & \relaccval{83.46} & 99.23 & \relaccval{81.54} & 99.15 & \relaccval{89.15} & 99.54 & \relaccval{76.78} & 99.48 & \relaccval{74.17} & 99.39 & \relaccval{67.09} & 99.43 \\
    \rowcolor{reverseRow}
    {} & 3-reverse & \relaccval{35.92} & 99.92 & \relaccval{43.67} & 99.75 & \relaccval{50.08} & 99.75 & \relaccval{52.08} & 99.83 & \relaccval{55.17} & 99.83 & \relaccval{58.08} & 99.67 & \relaccval{71.00} & 99.00 & \relaccval{72.08} & 99.08 & \relaccval{78.46} & 99.54 & \relaccval{5.57} & 99.09 & \relaccval{3.22} & 99.43 & \relaccval{2.78} & 99.57 \\
    \rowcolor{mixedRow}
    \multirow{-5}{*}{8B} & 3-mixed & \relaccval{27.17} & 99.67 & \relaccval{40.58} & 99.75 & \relaccval{51.50} & 99.75 & \relaccval{47.08} & 99.75 & \relaccval{38.25} & 99.75 & \relaccval{48.75} & 99.50 & \relaccval{86.08} & 98.77 & \relaccval{83.00} & 99.15 & \relaccval{92.08} & 99.00 & \relaccval{53.13} & 99.17 & \relaccval{47.96} & 99.43 & \relaccval{56.52} & 99.52 \\
    \cmidrule{1-26}
    \rowcolor{ablationRow}
    {} & prior-free & \relaccval{99.11} & 99.41 & \relaccval{99.11} & 99.11 & \relaccval{98.52} & 98.82 & \relaccval{99.46} & 99.46 & \relaccval{99.73} & 99.73 & \relaccval{99.20} & 99.20 & \relaccval{98.95} & 99.14 & \relaccval{99.23} & 99.33 & \relaccval{98.75} & 99.52 & \relaccval{98.64} & 98.86 & \relaccval{98.72} & 98.93 & \relaccval{99.29} & 99.43 \\
    \rowcolor{forwardRow}
    {} & prior-conform & \relaccval{99.83} & 99.83 & \relaccval{99.83} & 99.83 & \relaccval{99.75} & 99.75 & \relaccval{99.50} & 99.50 & \relaccval{99.50} & 99.58 & \relaccval{99.92} & 99.92 & \relaccval{99.23} & 99.46 & \relaccval{99.15} & 99.46 & \relaccval{99.15} & 99.31 & \relaccval{99.48} & 99.48 & \relaccval{99.39} & 99.39 & \relaccval{99.65} & 99.65 \\
    \rowcolor{reverseRow}
    {} & 2-reverse & \relaccval{83.67} & 99.75 & \relaccval{82.33} & 99.75 & \relaccval{84.17} & 99.83 & \relaccval{96.58} & 99.58 & \relaccval{95.67} & 99.67 & \relaccval{96.33} & 99.83 & \relaccval{91.46} & 99.23 & \relaccval{91.08} & 99.23 & \relaccval{90.38} & 99.54 & \relaccval{49.65} & 99.35 & \relaccval{49.13} & 99.26 & \relaccval{53.04} & 99.35 \\
    \rowcolor{reverseRow}
    {} & 3-reverse & \relaccval{20.08} & 99.83 & \relaccval{30.00} & 99.92 & \relaccval{33.08} & 99.83 & \relaccval{42.33} & 99.50 & \relaccval{60.92} & 99.58 & \relaccval{59.58} & 99.92 & \relaccval{41.23} & 99.15 & \relaccval{37.62} & 99.31 & \relaccval{36.62} & 99.62 & \relaccval{6.17} & 99.17 & \relaccval{6.22} & 99.04 & \relaccval{13.04} & 99.43 \\
    \rowcolor{mixedRow}
    \multirow{-5}{*}{32B} & 3-mixed & \relaccval{23.83} & 99.42 & \relaccval{39.50} & 99.50 & \relaccval{45.83} & 99.83 & \relaccval{48.75} & 99.33 & \relaccval{44.83} & 99.83 & \relaccval{46.33} & 100.00 & \relaccval{89.38} & 98.85 & \relaccval{91.23} & 99.15 & \relaccval{89.92} & 99.23 & \relaccval{38.00} & 98.91 & \relaccval{28.09} & 99.70 & \relaccval{36.00} & 99.65 \\
    \midrule
    \rowcolor{modelRow}
    \multicolumn{26}{c}{\textbf{GPT-5.4}} \\
    \cmidrule{1-26}
    \rowcolor{ablationRow}
    {} & prior-free & \relaccval{92.90} & 100.00 & \relaccval{99.41} & 100.00 & \relaccval{91.42} & 99.70 & \relaccval{96.25} & 99.73 & \relaccval{97.86} & 99.73 & \relaccval{75.34} & 100.00 & \relaccval{86.88} & 92.15 & \relaccval{97.89} & 99.23 & \relaccval{99.04} & 99.33 & \relaccval{98.93} & 99.07 & \relaccval{98.79} & 98.79 & \relaccval{99.86} & 99.86 \\
    \rowcolor{forwardRow}
    {} & prior-conform & \relaccval{98.00} & 99.67 & \relaccval{75.67} & 100.00 & \relaccval{90.67} & 100.00 & \relaccval{96.33} & 100.00 & \relaccval{62.33} & 99.83 & \relaccval{71.50} & 100.00 & \relaccval{97.38} & 98.92 & \relaccval{99.08} & 99.23 & \relaccval{98.77} & 99.08 & \relaccval{99.65} & 99.65 & \relaccval{99.48} & 99.48 & \relaccval{99.91} & 99.91 \\
    \rowcolor{reverseRow}
    {} & 2-reverse & \relaccval{89.00} & 99.67 & \relaccval{86.50} & 100.00 & \relaccval{90.67} & 100.00 & \relaccval{92.17} & 99.17 & \relaccval{68.17} & 100.00 & \relaccval{75.17} & 100.00 & \relaccval{96.92} & 98.62 & \relaccval{99.38} & 99.69 & \relaccval{99.54} & 99.54 & \relaccval{98.09} & 99.48 & \relaccval{98.17} & 99.57 & \relaccval{99.39} & 99.65 \\
    \rowcolor{reverseRow}
    {} & 3-reverse & \relaccval{86.00} & 99.50 & \relaccval{84.17} & 100.00 & \relaccval{94.83} & 100.00 & \relaccval{93.83} & 99.83 & \relaccval{78.17} & 100.00 & \relaccval{80.50} & 99.83 & \relaccval{76.31} & 98.00 & \relaccval{90.31} & 100.00 & \relaccval{98.62} & 100.00 & \relaccval{70.26} & 99.48 & \relaccval{86.61} & 99.48 & \relaccval{93.04} & 99.83 \\
    \rowcolor{mixedRow}
    \multirow{-5}{*}{5.4} & 3-mixed & \relaccval{92.33} & 99.50 & \relaccval{96.17} & 100.00 & \relaccval{98.33} & 99.50 & \relaccval{86.33} & 99.67 & \relaccval{96.00} & 100.00 & \relaccval{96.50} & 100.00 & \relaccval{86.31} & 99.23 & \relaccval{98.00} & 100.00 & \relaccval{99.38} & 99.54 & \relaccval{96.26} & 99.83 & \relaccval{99.04} & 99.48 & \relaccval{91.74} & 92.52 \\
    \cmidrule{1-26}
    \rowcolor{ablationRow}
    {} & prior-free & \relaccval{97.34} & 100.00 & \relaccval{100.00} & 100.00 & \relaccval{100.00} & 100.00 & \relaccval{83.11} & 100.00 & \relaccval{99.73} & 100.00 & \relaccval{98.93} & 100.00 & \relaccval{91.09} & 98.85 & \relaccval{92.05} & 99.43 & \relaccval{96.36} & 97.61 & \relaccval{97.36} & 99.50 & \relaccval{99.86} & 99.86 & \relaccval{99.64} & 99.64 \\
    \rowcolor{forwardRow}
    {} & prior-conform & \relaccval{98.42} & 99.50 & \relaccval{99.50} & 99.75 & \relaccval{100.00} & 100.00 & \relaccval{93.08} & 99.58 & \relaccval{98.83} & 99.75 & \relaccval{98.00} & 99.75 & \relaccval{93.38} & 99.38 & \relaccval{99.00} & 99.08 & \relaccval{96.77} & 99.54 & \relaccval{99.17} & 99.17 & \relaccval{99.74} & 99.74 & \relaccval{99.52} & 99.57 \\
    \rowcolor{reverseRow}
    {} & 2-reverse & \relaccval{71.25} & 99.00 & \relaccval{95.33} & 99.75 & \relaccval{92.75} & 100.00 & \relaccval{71.42} & 99.75 & \relaccval{93.58} & 99.92 & \relaccval{91.08} & 100.00 & \relaccval{83.92} & 99.46 & \relaccval{96.69} & 99.46 & \relaccval{95.08} & 99.77 & \relaccval{79.09} & 98.78 & \relaccval{90.35} & 99.70 & \relaccval{94.91} & 99.57 \\
    \rowcolor{reverseRow}
    {} & 3-reverse & \relaccval{47.42} & 99.92 & \relaccval{50.33} & 99.92 & \relaccval{53.75} & 100.00 & \relaccval{48.00} & 99.67 & \relaccval{57.08} & 100.00 & \relaccval{52.50} & 100.00 & \relaccval{86.69} & 99.00 & \relaccval{94.69} & 99.31 & \relaccval{95.92} & 99.38 & \relaccval{48.48} & 98.22 & \relaccval{70.30} & 99.70 & \relaccval{79.70} & 99.74 \\
    \rowcolor{mixedRow}
    \multirow{-5}{*}{5.4 Mini} & 3-mixed & \relaccval{48.75} & 99.58 & \relaccval{31.92} & 100.00 & \relaccval{35.42} & 100.00 & \relaccval{49.33} & 99.33 & \relaccval{47.25} & 99.92 & \relaccval{31.58} & 99.92 & \relaccval{86.00} & 99.23 & \relaccval{95.77} & 99.69 & \relaccval{95.00} & 99.54 & \relaccval{57.52} & 99.48 & \relaccval{52.74} & 99.57 & \relaccval{65.83} & 99.65 \\
    \bottomrule
  \end{tabular}%
  }
  \caption{Full per-cell results. Values are percentages for each image scale within each (model, condition, relation) cell. Conditions shown are \textit{prior-free}, \textit{prior-conform}, \textit{2-reverse}, \textit{3-reverse}, and \textit{3-mixed}. For each relation, the columns are grouped by image scale ($1\times$, $1.5\times$, $2\times$). \textbf{\textcolor{relaccRed}{RelAcc}} (= Relation Accuracy) and \textbf{EM} (= Class-name Exact Match) are reported under each scale. We use greedy decoding to ensure each cell has a deterministic single-sample accuracy.}
\label{tab:main_table}
\end{table*}%

\clearpage
\onecolumn
\begingroup
\setlength{\parindent}{0pt}
\thispagestyle{plain}

\begin{figure}[!ht]
\centering
\begin{tcolorbox}[
    enhanced,
    colback=white,
    colframe=black!35,
    boxrule=0.4pt,
    arc=4pt,
    left=8pt, right=8pt, top=8pt, bottom=8pt,
    title={\small System prompt},
    fonttitle=\bfseries,
    coltitle=white,
    colbacktitle=black!45!black,
    listing only,
    listing options={
        basicstyle=\ttfamily\scriptsize,
        breaklines=true,
        columns=fullflexible,
        keepspaces=true
    },
]
You are an accurate UML diagram reasoning assistant.
You will be asked to complete a question or task.
Treat the provided image as the only source of truth.
Do not use the model's internal knowledge, assumptions, and guesses.
Follow the requested output format exactly, without extra explanation.
\end{tcolorbox}
\caption{\textbf{System prompt.} General instruction shared by all evaluation prompts, emphasizing image-grounded reasoning and strict output formatting.}
\label{fig:prompt_system}
\end{figure}

\begin{figure}[!ht]
\centering
\begin{tcolorbox}[
    enhanced,
    colback=white,
    colframe=black!35,
    boxrule=0.4pt,
    arc=4pt,
    left=8pt, right=8pt, top=8pt, bottom=8pt,
    title={\small Stage-1 class-name recovery prompt},
    fonttitle=\bfseries,
    coltitle=white,
    colbacktitle=black!45!black,
    listing only,
    listing options={
        basicstyle=\ttfamily\scriptsize,
        breaklines=true,
        columns=fullflexible,
        keepspaces=true
    },
]
The image provided is a UML diagram showing \textbf{<Relation>} relationships.
The UML diagram contains \textbf{<Number>} classes.
Each class is represented as a box, with the class name at the top.
The green circled letter 'C' inside each box is only a class-type icon and is not part of the class name. The actual class name is the text to the right of the icon.
You must treat the image as the only source of truth for the task.
Do not infer, guess, or invent any names. Complete the following task using only the class names that are visibly and explicitly depicted in the UML diagram.

The task is:
List all class names that appear in the UML diagram.

Each class name in your final output must match the UML diagram text exactly, character by character.
You may reason privately, but do not reveal any reasoning or intermediate steps.
Output only a JSON array of strings, for example: ["ClassA", "ClassB"].
Do not output any text before or after the JSON array.
Output [] if the image is missing or unreadable.
\end{tcolorbox}
\caption{\textbf{Class-name recovery prompt.} Prompt used to recover the class names explicitly visible in the UML diagram before relation-direction evaluation.}
\label{fig:prompt_stage1}
\end{figure}

\begin{figure}[!ht]
\centering
\begin{tcolorbox}[
    enhanced,
    colback=white,
    colframe=black!35,
    boxrule=0.4pt,
    arc=4pt,
    left=8pt, right=8pt, top=8pt, bottom=8pt,
    title={\small Stage-2 relation-direction QA prompt},
    fonttitle=\bfseries,
    coltitle=white,
    colbacktitle=black!45!black,
    listing only,
    listing options={
        basicstyle=\ttfamily\scriptsize,
        breaklines=true,
        columns=fullflexible,
        keepspaces=true
    },
]
The image provided is a UML diagram showing \textbf{<Relation>} relationships.
The diagram contains \textbf{<Two/Three>} classes: \textbf{<Class names>}.
Each class is represented as a box, with the class name at the top.
The green circled letter 'C' inside each box is only a class-type icon and is not part of the class name. The actual class name is the text to the right of the icon.

\textbf{<Relation-specific UML rule>}

Carefully verify the direction of every arrow or line in the UML diagram. Treat arrow direction as authoritative and do not assume it.
You must treat the image as the only source of truth for the question.
Do not use the model's internal knowledge or prior assumptions to attempt to answer the question.
Answer the following question solely from the relationships explicitly depicted in the image.

The question is: \textbf{<Relation-specific question templates>}

You may reason privately, but do not reveal any reasoning or intermediate steps. Output only one of: True, False, or Unknown.
Output 'Unknown' if the image is missing, unreadable, or the relation direction is unclear.
\end{tcolorbox}
\caption{\textbf{Relation-direction QA prompt.} Prompt used to answer the relation-direction question after the class names have been specified.}
\label{fig:prompt_stage2}
\end{figure}

\clearpage

\begin{figure}[!ht]
\centering
\begin{tcolorbox}[
    enhanced,
    colback=white,
    colframe=black!35,
    boxrule=0.4pt,
    arc=4pt,
    left=8pt, right=8pt, top=8pt, bottom=8pt,
    title={\small Relation-specific rules and query templates},
    fonttitle=\bfseries,
    coltitle=white,
    colbacktitle=black!45!black,
    listing only,
    listing options={
        basicstyle=\ttfamily\scriptsize,
        breaklines=true,
        columns=fullflexible,
        keepspaces=true
    },
]
Relation-specific UML rules:

Inheritance: A solid line with a hollow triangular arrow indicates an inheritance relationship. The arrow points from the subclass to the superclass.

Aggregation: A solid line with a hollow diamond indicates an aggregation relationship. The class at the diamond end is the whole, and the class at the other end is the part.

Composition: A solid line with a filled diamond indicates a composition relationship. The class at the diamond end is the whole, and the class at the other end is the part.

Dependency: A dashed line with an open arrow indicates a dependency relationship. The arrow points from the dependent class to the class it depends on.
\medskip
\medskip

Relation-specific question templates:

Inheritance: Does class \textbf{<ClassA>} inherit from class \textbf{<ClassB>}?

Aggregation: Is class \textbf{<ClassA>} the whole and class \textbf{<ClassB>} the part?

Composition: Is class \textbf{<ClassA>} the whole and class \textbf{<ClassB>} the part?

Dependency: Does class \textbf{<ClassA>} depend on class \textbf{<ClassB>}?
\end{tcolorbox}
\caption{\textbf{Relation-specific UML rules and query templates.} UML notation rules and question templates used to instantiate the relation-direction QA prompt.}
\label{fig:prompt_rules_templates}
\end{figure}
\endgroup

\end{document}